%% file: main.tex
\documentclass[10pt,twocolumn,letterpaper]{article}

\usepackage{cvpr}              


\usepackage{url}

\usepackage{xspace}
\usepackage{dsfont}
\usepackage{afterpage}

\usepackage{enumitem}
\usepackage{booktabs}
\usepackage{arydshln}
\usepackage{caption}
\usepackage{caption} 
\usepackage{multirow} 
\usepackage{enumitem}
\usepackage{colortbl} 
\usepackage{bbm}
\usepackage{algorithm}
\usepackage{algpseudocode}                  
\usepackage{setspace}                       
\usepackage{lipsum} 
\usepackage{subcaption} 
\usepackage{xcolor}
\usepackage{pifont}
\newcommand{\cmark}{\textcolor{red}{\ding{51}}}   
\newcommand{\xmark}{\textcolor{cyan}{\ding{55}}}  
\usepackage{wrapfig}
\usepackage{graphicx}
\usepackage{caption}

\usepackage{mathtools} 
\usepackage{bm} 
\usepackage{soul} 
\usepackage{nicefrac}
\usepackage{csquotes}
\usepackage{array}
\usepackage{duckuments}
\definecolor{cvprblue}{rgb}{0.21,0.49,0.74}

\usepackage[pagebackref,breaklinks,colorlinks,allcolors=cvprblue]{hyperref}

\def\paperID{} 
\def\confName{CVPR}
\def\confYear{2026}

\title{\textbf{RefTon}: \textbf{Ref}erence person shot assist virtual \textbf{T}ry-\textbf{on}}

\author{
Liuzhuozheng Li$^{1,2}$\thanks{Equal contribution.} ~
Yue Gong$^{2*}$ ~
Shanyuan Liu$^{2*}$ ~
Zanyi Wang$^{3}$ ~
Dengyang Jiang$^{4}$ ~ \\
Leibucha Wu$^{2}$ ~
Bo Cheng$^{2}$ ~
Yuhang Ma$^{2}$ ~
Dawei Leng$^{2}$\thanks{Corresponding author} ~
Yuhui Yin$^{2}$ \\[2mm]
\fontsize{10.4pt}{9.84pt}\selectfont
$^{1}$ The University of Tokyo \hspace{6mm}
$^{2}$ 360 AI Research \hspace{6mm}
$^{3}$ University of California, San Diego \\
$^{4}$ The Hong Kong University of Science and Technology \\
\textit{Code is available at:} \url{https://github.com/360CVGroup/RefTon}.
}

\begin{document}
\definecolor{lightgreen}{RGB}{223,243,212}
\pagenumbering{arabic}

\twocolumn[{%
\renewcommand\twocolumn[1][]{#1}%
\maketitle

\begin{center}
    \vspace{-0.3in}
    \includegraphics[width=0.8\textwidth]{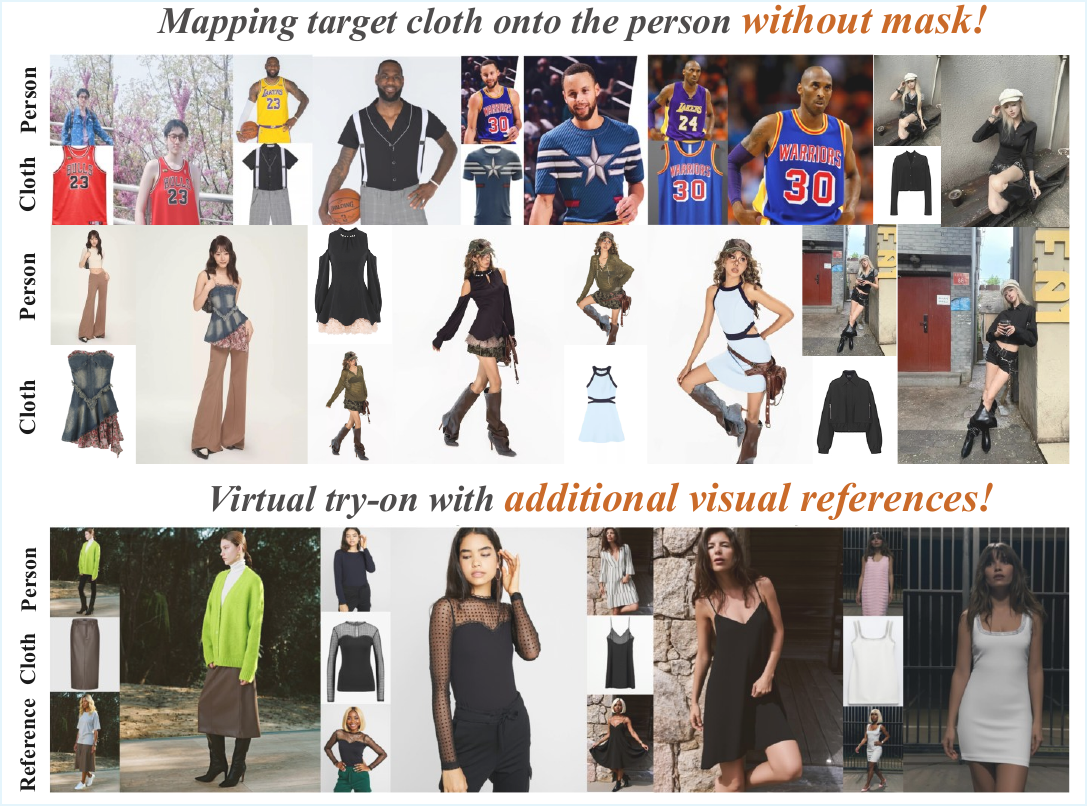}
    \captionof{figure}{\textbf{In-the-wild try-on results produced by our RefTON model.}
    The first row demonstrates our \textbf{mask-free try-on} capability, where the garment is transferred directly to the target person. The second row shows our \textbf{additional-reference try-on} mode, in which extra visual references are incorporated to enhance structural accuracy, texture fidelity, and overall realism.}
    \label{fig:in-the-wild}
\end{center}
}]
\begingroup
\renewcommand\thefootnote{\fnsymbol{footnote}}
\footnotetext[1]{Equal contribution.}
\footnotetext[2]{Corresponding author}
\endgroup

\input{sec/0_abstract}
\input{sec/1_intro}
\input{sec/2_related_work}

\input{sec/3_method}
\input{sec/4_experiment}

\section{Acknowledgments}
This research was supported by 360 AI Research. We sincerely thank 360 AI Research for its support of this work.

{
    \small
    \bibliographystyle{ieeenat_fullname}
    \bibliography{main}
}

\input{sec/5_supplentary}
\end{document}

%% file: sec/0_abstract.tex
\begin{abstract}
We introduce RefTon, a flux-based person-to-person virtual try-on framework that enhances garment realism through unpaired visual references. Unlike conventional approaches that rely on complex auxiliary inputs such as body parsing and warped mask or require finely designed extract branches to process various input conditions, RefTon streamlines the process by directly generating try-on results from a source image and a target garment, without the need for structural guidance or auxiliary components to handle diverse inputs. Moreover, inspired by human clothing selection behavior, RefTon leverages additional reference images (the target garment worn on different individuals) to provide powerful guidance for refining texture alignment and maintaining the garment details. To enable this capability, we built a dataset containing unpaired reference images for training. Extensive experiments on public benchmarks demonstrate that RefTon achieves competitive or superior performance compared to state-of-the-art methods, while maintaining a simple and efficient person-to-person design.
\end{abstract}

%% file: sec/1_intro.tex
\section{Introduction}
The \textbf{Virtual Try-On (ViTON)} model aims to generate photo-realistic images of a person wearing target clothing, a tool crucial for applications in online retail and personalized fashion systems. ViTON methods are broadly categorized into Generative Adversarial Networks (GANs) \cite{goodfellow2020generative} and Diffusion Models \cite{Ho2020DenoisingDP, Rombach2021HighResolutionIS}. Early ViTON research relied on GANs \cite{choi2021viton, han2018viton, wang2018toward}, which typically employed warping modules to deform clothing for alignment with the human body, followed by fusion to achieve visual harmony. However, GAN-based approaches frequently generate unrealistic artifacts, particularly when dealing with complex clothing textures or challenging human poses. Recently, methods based on latent diffusion models (LDMs) \cite{chen2024wear, xu2025ootdiffusion} have gained traction, significantly enhancing clothing warping and addressing structural arrangement and texture preservation during denoising \cite{kim2024stableviton, zhu2023tryondiffusion, xu2024magicanimate, choi2024improving}. Despite these advances, current diffusion-based ViTON technologies still generally rely on extensive auxiliary conditions, such as clothing region masks, garment masks, human poses, key points, or multi-modal inputs like text prompts \cite{xu2025ootdiffusion, kim2024promptdresser, chong2024catvton, chong2025catv2ton, yang2025omnivton}.

\begin{figure}[t]
    \centering
    \includegraphics[width=\linewidth]{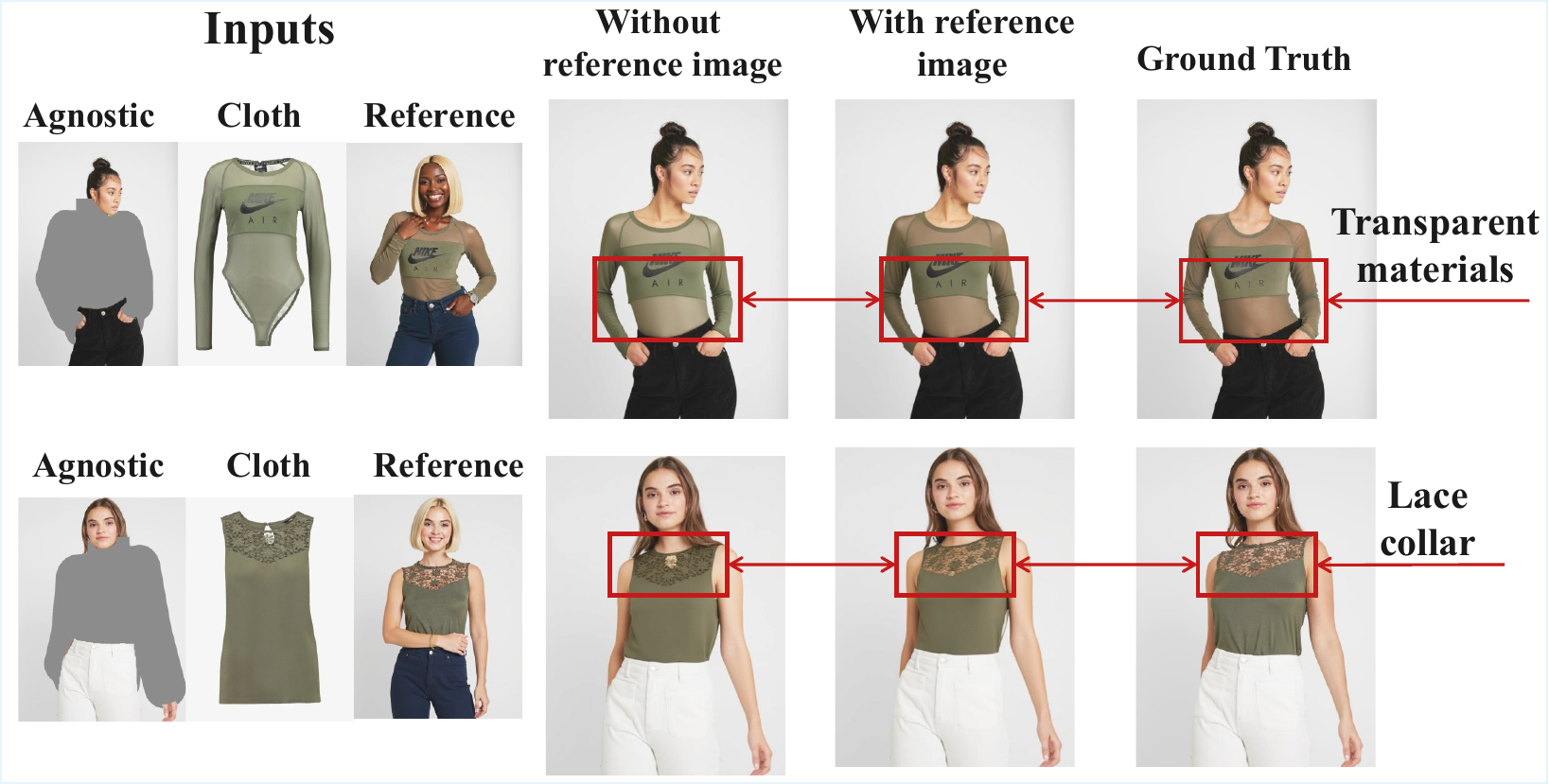}
    \caption{\textbf{Effect of reference images in virtual try-on.} From left to right: inputs (agnostic, cloth, reference), results without reference images, results with reference images, and ground truth. Using reference images in both training and inference improves visual fidelity and preserves fine details (e.g., transparent materials and lace collars). Zoom in for a better comparison.}
    \label{Fig:reference_effect}
\end{figure}

Despite the remarkable progress of prior virtual try-on approaches, they are still constrained by two critical limitations that hinder the authenticity of the try-on results and broader applicability:
 \textbf{First}, these approaches rely on multiple external models and internal modules, such as pose estimators~\cite{guler2018densepose, toshev2014deeppose, openpose, cao2017realtime, wei2016cpm}, human parse models~\cite{li2020self, dong2014towards}, segmentation models~\cite {kirillov2023segment, ravi2024sam2}, to process different conditions, which compromises the practicality. To process diverse inputs, additional modules are integrated into the model, which consequently increases the overall framework complexity. Moreover, in practical applications, the quality of conditional inputs---such as the cloth mask---has a substantial impact on the quality of the final try-on results;
\textbf{Second}, many aspects of clothing, such as style, texture, and detailed design, cannot be fully perceived from the garment image alone; instead, it is more important to consider the overall appearance when a model wears the garment. Therefore, in real-world try-on scenarios, such as online shopping, users are typically more interested in model images rather than the garment itself. They tend to see how the target garment looks when worn on a real person, rather than relying solely on the isolated garment image as a reference. For example, as illustrated in Fig.~\ref{Fig:reference_effect}, at the garment in the first row, it is difficult to tell whether it is a green translucent fabric or a light green opaque one, whereas the reference image clearly reveals its green translucent material. We cannot accurately identify transparent materials or intricate designs, such as lace collars, solely from cloth images. In contrast, the reference images of human models wearing the garments reveal such details. However, existing virtual try-on methods do not support such references due to the lack of corresponding reference data in public datasets~\cite{morelli2022dress, choi2021viton, fang2024vivid, zheng2019virtually, shen2025imagdressing, he2024wildvidfit, meng2025hf}.

Based on the above observation, we propose \textbf{RefTon}, a flux-based person-to-person virtual try-on framework that achieves strong performance \textit{without relying on any external models or auxiliary components}, while being further enhanced by \textit{additional reference images} that offer more accurate and context-aware guidance for the try-on model. First, to ensure the best performance, we adopt the powerful image editing model \textit{Flux-kontext} as our base and apply adaptation on its position index for RoPE~\cite{su2024roformer} to make it suitable for multi-condition/resolution virtual try-on inputs. Similar to~\cite{chong2024catvton}, RefTon eliminates the need for auxiliary inputs such as segmentation masks using a two-stage training strategy, allowing for simple inference with only a source image and the target garment as inputs. Second, we introduce the use of images of a different person clothed in the target garment as the visual references, like the model images in online shopping, which better reflect users' real-world behavior when choosing clothes and enable the preservation of fine garment details that existing methods cannot achieve. To achieve these objectives, we propose a reference data generation pipeline, by which we construct a dataset with supplementary reference images and use unpaired person-cloth samples to train our own model to utilize reference images as additional visual guidance. These improvements empower RefTon to achieve both a simplified model structure and a streamlined inference process, while simultaneously delivering superior generation results.

In summary, the main contributions are as follows:
\begin{itemize}
\item We propose incorporating \textbf{additional reference images} into the virtual try-on pipeline. This significantly enhances the authenticity and visual quality of the try-on results, achieving \textbf{State-of-the-Art} performance in preserving fine garment design details.

\item We designed a \textbf{reference data generation framework} to create the necessary reference images for both the clothing and target ground truth samples. Based on this pipeline, we built the VFR dataset upon existing benchmarks (e.g., VITON-HD, DressCode, ViViD), providing a robust new resource to improve the practicality and evaluation of virtual try-on models.

\item We present an adaptation of the \textit{Flux-Kontext} I2I model with a modified \textit{Rescaled Position Indexing} mechanism to support \textbf{flexible multi-conditional and multi-resolution inputs}, along with a two-stage training strategy for virtual try-on. Our framework enables integration of varying numbers and types of reference images and effectively supports mask-based and person-to-person try-on within a single model. It achieves \textbf{state-of-the-art} performance and demonstrates strong generalization to in-the-wild person--clothing scenarios.

\end{itemize}

%% file: sec/2_related_work.tex
\section{Related Works}
\subsection{Generative Model via Flow Matching}
Generative modeling has rapidly progressed with diffusion models (DMs)~\cite{sohl2015deep}, score-based generative models (SGMs)~\cite{song2020score}, and flow-based methods. Recent work has explored controllable generation, layout-conditioned synthesis, unified planning-generation frameworks~\cite{cheng2024hico, he2025plangen}, as well as diffusion transformer variants for multilingual generation, efficient sampling, and fine-grained control~\cite{Liu_2025, ma2025nami, liu2025nanocontrollightweightframeworkprecise, gong2025cta}. Flow-based methods~\cite{dinh2016density, dinh2014nice, kingma2018glow} have been further advanced to address the inefficiency of continuous normalizing flows (CNFs), which require costly backpropagation through ODE solvers during training~\cite{Chen2018NeuralOD}. Flow Matching (FM)~\cite{Lipman2022FlowMF} mitigates this limitation by learning a time-dependent vector field that deterministically transports a simple before the data distribution, using a simulation-free objective that avoids numerical integration during training. By directly parameterizing the probability flow, FM achieves competitive or superior sample quality compared to diffusion models with significantly fewer sampling steps. Our method builds upon the \textit{Flux-Kontext} architecture, where input images are encoded into latent representations, flattened into sequences, and concatenated with Gaussian noise $\boldsymbol{\epsilon}$. It is also closely related to recent FLUX-based appearance editing and transfer approaches~\cite{zhu2025flux}.

\subsection{Diffusion-based Virtual Try-on}
Diffusion models~\cite{Ho2020DenoisingDP, Rombach2021HighResolutionIS} have enabled significant progress in garment or makeup transfer~\cite{kim2024stableviton, zhu2023tryondiffusion, choi2021viton, zhu2025flux}. Leveraging the flexibility of Stable Diffusion~\cite{morelli2023ladi, kim2024stableviton}, prior works exploit text guidance and inpainting for garment synthesis. Extensions such as DiffusionCLIP~\cite{kim2022diffusionclip} introduce semantic control via CLIP, while methods like DCI-VTON~\cite{gou2023taming} and IDM-VTON~\cite{choi2021viton} adopt two-stage pipelines to align and fuse garments, improving structural consistency.

Recent approaches, including CatVTON~\cite{chong2024catvton}, OmniTry~\cite{feng2025omnitry}, Any2AnyTryon~\cite{guo2025any2anytryon}, and OmniVTON~\cite{yang2025omnivton}, explore person-to-person try-on without explicit masks. However, these methods typically rely on additional conditions (e.g., pose) or fail to unify mask-based and mask-free settings. Moreover, P2P pipelines such as TryOffDiff~\cite{velioglu2025enhancing} and ViTON-GUN~\cite{ViTON-GUN} adopt a ``try-off--then--try-on'' strategy, which introduces error accumulation and loss of garment details. In contrast, our method directly leverages the reference image, avoiding the try-off stage and better preserving garment structure and material fidelity. In summary, existing diffusion-based virtual try-on methods either rely on heavy auxiliary annotations or lack support for clothed reference images. To address these limitations, we propose RefTON, which adapts the virtual try-on task to the Flux-Kontext framework, enabling end-to-end, reference-guided generation while supporting both mask-based inpainting and mask-free editing.

%% file: sec/3_method.tex
\begin{figure*}[t]
    \centering
    \includegraphics[width=0.75\linewidth]{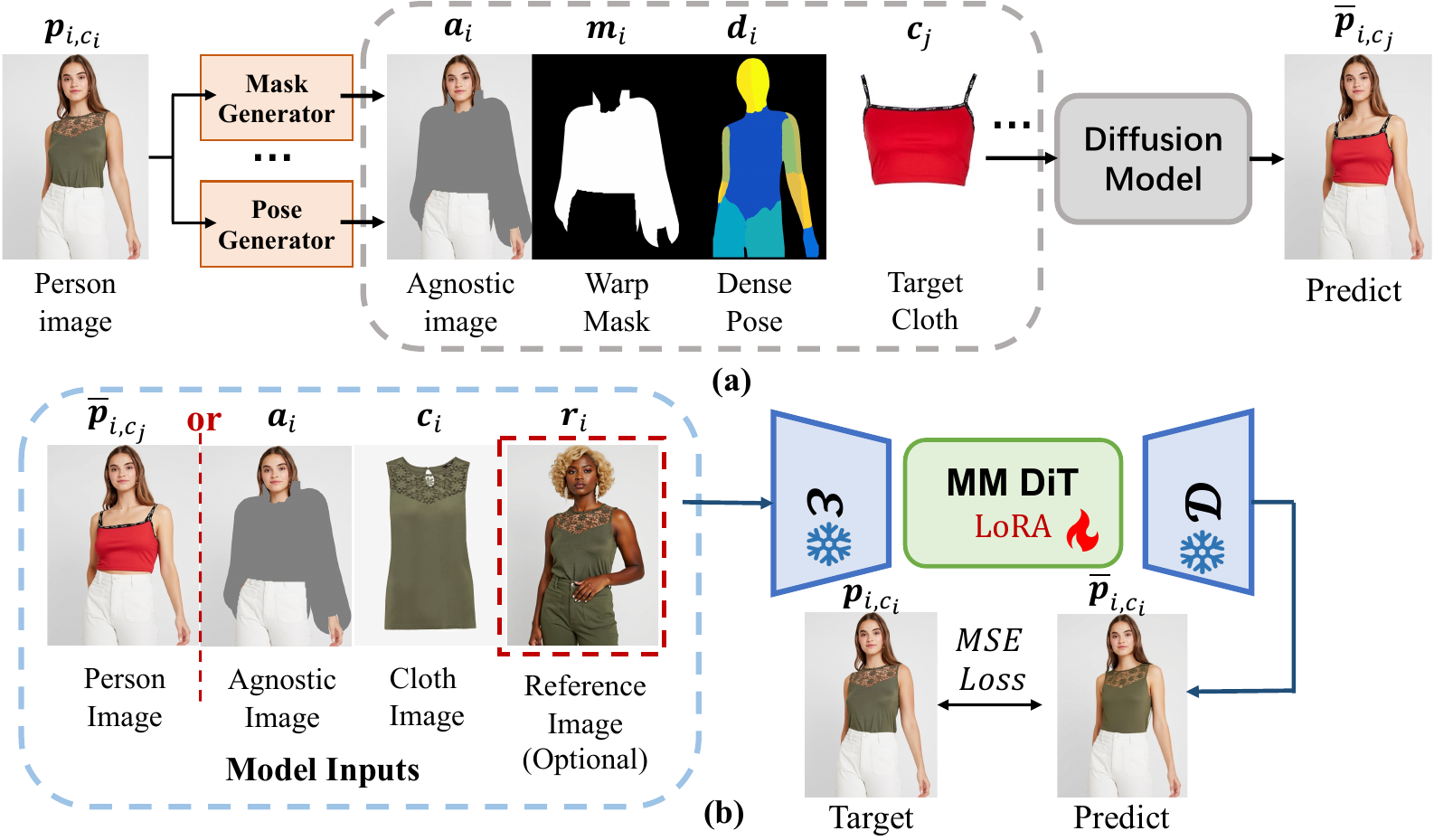}
    \caption{Overview of our two-stage training strategy. (a) Stage 1 follows mask-based try-on paradigms to generate person images wearing random garments from masked inputs, providing training data for Stage 2. (b) Stage 2 uses synthesized person images, along with the target garment and optional reference images, to train a person-to-person virtual try-on model that directly fits the garment onto the person.}
    \label{Fig:two_stage}
\end{figure*}

\section{Method}
\subsection{Preliminary}
RefTon is built upon DiT~\cite{peebles2023scalable}, a scalable Transformer architecture for diffusion-based generation. Images are encoded into a latent space via an autoencoder~\cite{kingma2013auto} and then patched into tokens~\cite{dosovitskiy2020image}. The diffusion process~\cite{Ho2020DenoisingDP} operates on these tokens, with the Transformer consuming noisy tokens and predicting their denoised results.

We consider the problem of generating images under the condition $\boldsymbol{y}$, which may represent garment images, semantic maps, human pose, or other modality-specific control signals. Let $\boldsymbol{x}$ denote the latent image representation obtained from a VAE encoder. The goal of \textit{Flux.1}~\cite{flux2024, labs2025flux1kontextflowmatching} is to approximate the conditional distribution $p(\boldsymbol{x}\mid\boldsymbol{y})$ by learning a time-dependent velocity field $\boldsymbol{v}(\boldsymbol{x}, \boldsymbol{y}, t)$ that transports a sample from a simple prior $\mathcal{N}(\boldsymbol{x};\boldsymbol{0}, \boldsymbol{I})$ at $t=0$ to the data distribution $p_{\text{data}}(\boldsymbol{x}|\boldsymbol{y})$ at $t=1$. The dynamics of the conditional probability density $p(\boldsymbol{x}|\boldsymbol{y}, t)$ over time $t$ are governed by the continuity equation:
\begin{equation}
\begin{aligned}
    \frac{\partial}{\partial t} p(\boldsymbol{x}|\boldsymbol{y}, t) 
    &= - \nabla_{\boldsymbol{x}} \cdot \big( 
    \boldsymbol{v}(\boldsymbol{x}, \boldsymbol{y}, t) \cdot p(\boldsymbol{x}|\boldsymbol{y}, t)
    \big), \quad 
    \\ &\boldsymbol{x}_{t=0} \sim \mathcal{N}(\boldsymbol{x};\boldsymbol{0}, \boldsymbol{I}),\ \boldsymbol{x}_{t=1} \sim p_{\text{data}}.
\end{aligned}
    \label{eq:continuity}
\end{equation}
To estimate $\boldsymbol{v}(\boldsymbol{x}, \boldsymbol{y}, t)$, we train a diffusion-transformer backbone to approximate the neural velocity field $\boldsymbol{v}_{\boldsymbol{\theta}}$ using the conditional flow matching objective~\cite{Lipman2022FlowMF, liu2022flow}:
\begin{equation}
\begin{aligned}
    \mathcal{L}_{\boldsymbol{\theta}}
    &= 
    \mathbb{E}_{t, \boldsymbol{x}_i, \boldsymbol{\epsilon}, \boldsymbol{y}_i}
    \Big[
    \big\| 
    \boldsymbol{v}_{\boldsymbol{\theta}}(\boldsymbol{x}, \boldsymbol{y}_i, t)
    - (\boldsymbol{x}_i - \boldsymbol{\epsilon})
    \big\|_2^2
    \Big], \\ \quad 
    &\boldsymbol{x} = (1 - t)\,\boldsymbol{x}_i + t\,\boldsymbol{\epsilon},
\end{aligned}
    \label{eq:loss}
\end{equation}
where $t \sim \mathcal{U}(0,1)$, $\boldsymbol{x}_i \sim \mathcal{X}_{\mathrm{train}}$, and $\boldsymbol{\epsilon} \sim \mathcal{N}(\boldsymbol{0}, \boldsymbol{I})$. This training objective encourages the model to learn a velocity field $\boldsymbol{v}_{\boldsymbol{\theta}}(\boldsymbol{x}, \boldsymbol{y}, t)$ that consistently guides the noisy samples toward the data distribution conditioned on $\boldsymbol{y}$, following the probability flow ODE starting from the Gaussian prior:
\begin{equation}
    d\boldsymbol{x} = \boldsymbol{v}(\boldsymbol{x}, \boldsymbol{y}, t)dt,
    \label{eq:infer}
\end{equation}
enabling controllable image synthesis at inference time.

In the virtual try-on setting, let $\boldsymbol{x}_i$ denote the image of a person wearing the target cloth, and let $\boldsymbol{y}_i$ represent a collection of conditional inputs, including the cloth-agnostic image $\boldsymbol{a}_i$, the target cloth $\boldsymbol{c}_i$, the visual references $\boldsymbol{r}_i$, and others. Formally, we write 
$\boldsymbol{y}_i = [\boldsymbol{a}_i, \boldsymbol{c}_i, \dots].$
The objective is to progressively transform a Gaussian noise sample $\boldsymbol{\epsilon}$ into the target image $\boldsymbol{x}_i$ guided by conditions $\boldsymbol{y}_i$.

\subsection{Person To Person Virtual Try-on Model with Two Stage Training}

Our goal is to dress the person directly with the target garment, without relying on auxiliary conditions such as DensePose~\cite{guler2018densepose} or segmentation masks. To achieve this, we train a diffusion model on clothing-person pairs $(\boldsymbol{c}_i, \boldsymbol{\bar p}_{i,\boldsymbol{c}_j})$ to generate the target image $\boldsymbol{p}_{i,\boldsymbol{c}_i}$, where the person wears the target garment, as shown in Fig.~\ref{Fig:two_stage}(b). This setup requires unpaired triplets $[\boldsymbol{\bar p}_{i,\boldsymbol{c}_j}, \boldsymbol{c}_i, \boldsymbol{p}_{i,\boldsymbol{c}_i}]$, where $\boldsymbol{c}_j \neq \boldsymbol{c}_i$. However, existing open-source benchmarks typically provide only paired data $[\boldsymbol{c}_i, \boldsymbol{p}_{i,\boldsymbol{c}_i}]$. Following the two-stage training strategy of CATVTON~\cite{chong2024catvton}, we adopt a similar pipeline for our try-on model and further exploit richer conditions, including agnostic masks and DensePose, to improve unpaired image generation. Specifically, in the first stage of RefTon training, we synthesize unpaired person images $\boldsymbol{\bar p}_{i,\boldsymbol{c}_j}$ using a mask-based try-on model.

As illustrated in Fig.~\ref{Fig:two_stage}(a), we train a virtual try-on model using agnostic person images $\boldsymbol{a}_i$, clothing images $\boldsymbol{c}_i$, densepose maps $\boldsymbol{d}_i$, warp masks $\boldsymbol{m}_i$ as inputs. During inference, a random unpaired garment $\boldsymbol{c}_j$ is selected to generate the corresponding synthesized person image $\boldsymbol{\bar p}_{i,c_j}$.
To ensure the quality of the synthesized person image $\boldsymbol{\bar p}_{i,c_j}$, the agnostic–cloth pairs $[\boldsymbol{a}_i, \boldsymbol{c}_i]$ used for training must belong to the same garment category (e.g., if $\boldsymbol{a}_i$ is from the “dresses” subset, the selected garment $\boldsymbol{c}_j$ should also come from “dresses” rather than “upper body” or “lower body”). Otherwise, the generated person image may appear unrealistic due to mismatches between the clothing mask region and the target garment (e.g., fitting a skirt onto the upper body or a shirt onto the lower body).

After obtaining the unpaired person images, we train the person-to-person model with either the agnostic image $\boldsymbol{a}_i$ or the generated person image $\boldsymbol{\bar p}_{i,c_j}$ from the first stage, sampled with equal $50\%$ probability. Instead of training a try-on model from scratch, we freeze the encoder and decoder of \textit{Flux-kontext}~\cite{labs2025flux1kontextflowmatching} and fine-tune only the transformer blocks using Low-Rank Adaptation~\cite{hu2022lora}. The model is optimized with the flow-matching objective in equation.~\ref{eq:loss}. In addition, to help the model capture garment appearance from another person, we provide an extra reference image $\boldsymbol{r}_i$ with a probability of $25\%$, as illustrated in Fig.~\ref{Fig:two_stage}(b).

\begin{figure}[t]
    \centering
    \includegraphics[width=\linewidth]{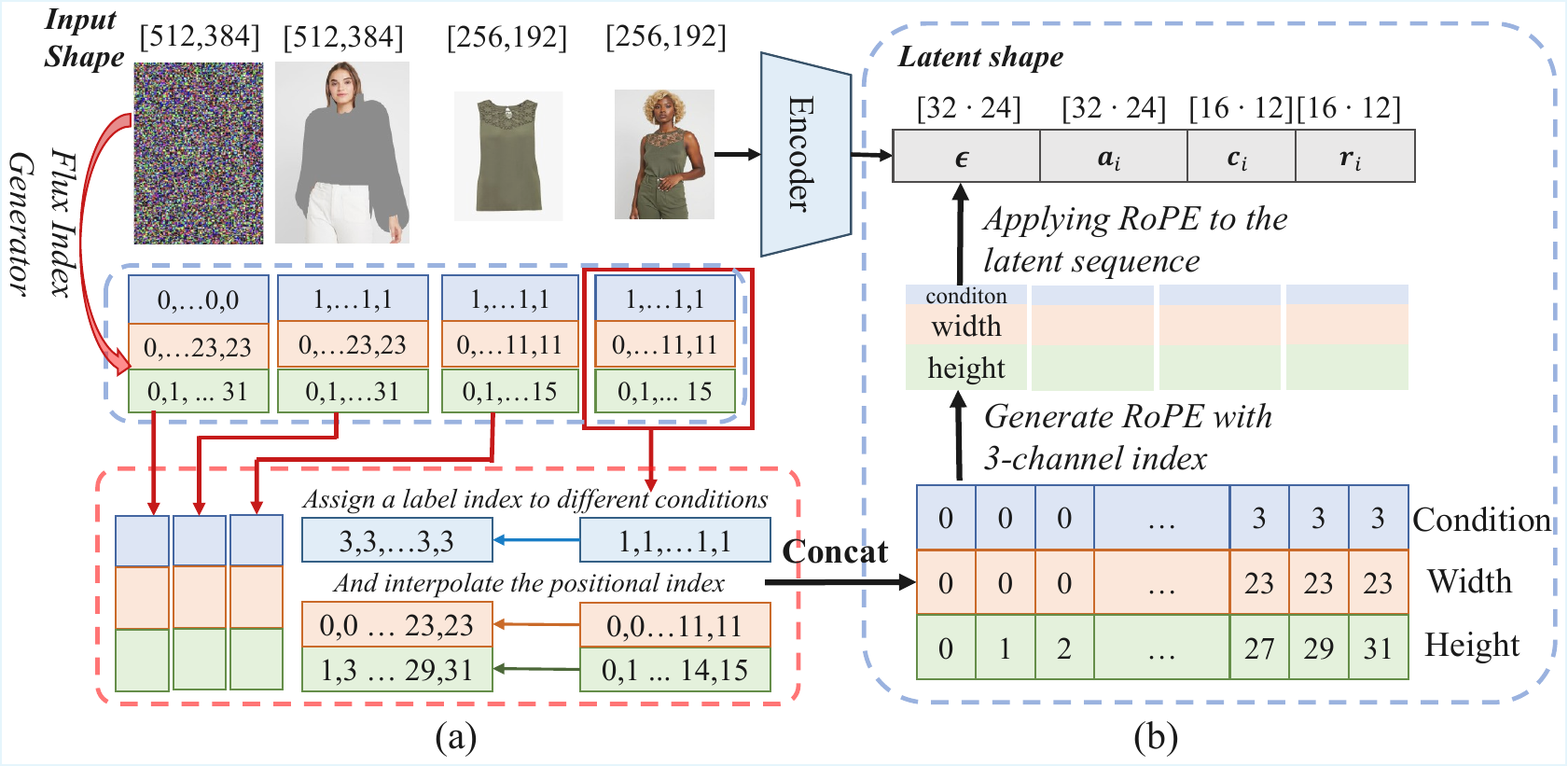}
    \caption{\textbf{Rescaled three-channel position index.} (a) The first channel encodes conditional inputs, while the second and third capture spatial positions for varying resolutions. (b) The concatenated indices are used to generate $RoPE$, which is applied to the latent sequence in the attention mechanism.}
    \label{Fig:index_generation}
\end{figure}

\subsection{Multi-input Training and Adaptation}

The latent embeddings are concatenated into the sequence $[\boldsymbol{\epsilon}_i, \boldsymbol{a}_i, \boldsymbol{c}_i]$ after image encoding in vanilla \textit{Flux-kontext}, and a index generator produces a position index of shape $[L,3]$, which is further converted into Rotary Position Embeddings (RoPE) for self-attention. Its first channel is a binary mask that distinguishes Gaussian noise $\boldsymbol{\epsilon}_i$ from conditional image/text inputs, while the second and third channels encode horizontal and vertical coordinates, respectively (Fig.~\ref{Fig:index_generation}). However, this binary design is insufficient for our setting, where the model must handle multiple heterogeneous image conditions. Therefore, we extend the first channel from a binary flag to discrete condition labels, allowing the transformer to distinguish different input types such as person, garment, and reference images.

Specifically, our model follows the \textit{Flux-kontext} to encode each image condition independently into latent patches, and concatenates to form a sequence such as $[\boldsymbol{a}_i, \boldsymbol{c}_i, \boldsymbol{r}_i, \ldots]$, which is then concatenated with the noise latent $\boldsymbol{\epsilon}_i$ along the sequence dimension. For each condition, we generate an individual three-channel position index: the first channel indicates the condition identity, and the second and third channels store integer spatial coordinates rescaled by the resolution ratio between the target image and the corresponding condition image, which preserves spatial alignment across different resolutions. This \textbf{\textit{Rescaled Position Index}} design enables flexible integration of multi-type and multi-resolution image conditions within a unified DiT framework. Similar indexing strategies have been explored in prior work~\cite{guo2025any2anytryon}, where positional indices are constructed on a concatenated image canvas, placing different conditions in a shared spatial layout. In contrast, our method assigns positional indices to each condition independently rather than generating them from a pixel-space concatenated canvas, resulting in a more flexible formulation for multi-resolution inputs. The modified DiT position indexing scheme is illustrated in Fig.~\ref{Fig:index_generation}.


\begin{figure*}[t]
  \centering
  \includegraphics[width=0.99\linewidth]{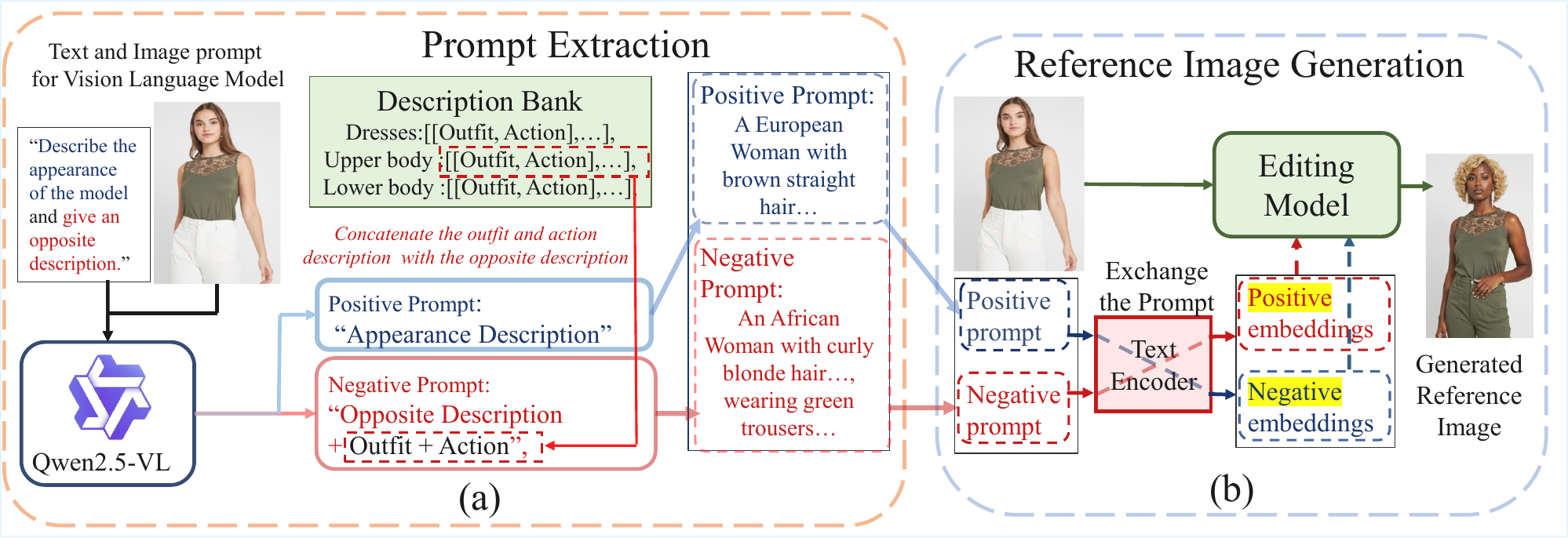}
  \caption{
Reference image generation pipeline. Given a target image of a European woman wearing the target upper-body garment, \textit{Qwen2.5-VL}~\cite{bai2025qwen2} produces an appearance description (e.g., "a European woman with brown straight hair") and an opposite description (e.g., "an African woman with curly blonde hair"). In (a), the opposite description is combined with the action and non-target clothing attributes to form the positive prompt, while the appearance description is used as the negative prompt. In (b), the image and prompts are fed into \textit{Flux-kontext} to generate reference images of different individuals wearing the target garment.
}
  \label{Fig:qwen}
\end{figure*}

\subsection{Virtual-Tryon Generation with Extra Visual Reference}
\label{section:dataset}

For both humans and generative virtual try-on models, garment images and conditions extracted by external models alone are insufficient to capture the realistic visual effect of wearing a target garment, as illustrated in Fig.~\ref{Fig:reference_effect}. A garment’s style, texture, and fine design details are more faithfully presented when worn by another person than when shown in isolation. Therefore, we introduce reference person images $\boldsymbol{r}_i$, in which another person wears the target garment, to provide more intuitive visual guidance during virtual try-on generation. To obtain such references, we construct pairs of the form $[\boldsymbol{c}_i, \boldsymbol{r}_i]$, representing "different persons wearing the target garment." Since these reference images $\boldsymbol{r}_i$ are unavailable in existing open-source virtual try-on datasets, we develop a reference data generation pipeline to synthesize them. This pipeline augments current datasets with supplementary references and enables model training with additional visual guidance.

Existing open-source datasets like VITON-HD~\cite{choi2021viton}, DressCode~\cite{morelli2022dress}, and IGPairs~\cite{shen2025imagdressing}, lack unpaired reference images $\boldsymbol{r}_i$ showing \emph{different persons wearing the target garment}. To overcome this limitation, we employ editing models to synthesize such reference images. To serve as accurate and informative visual guidance, the generated reference data $\boldsymbol{r}_i$ should satisfy the following requirements:

\begin{itemize}
    \item \textbf{Preserve the target garment faithfully.} The target garment's color, texture, and design must remain unchanged to ensure an accurate reference.

    \item \textbf{Introduce diversity in the person's appearance.} The person wearing the target garment in the reference image should be different from the target person wearing the same garment. Otherwise, the model may learn shortcuts and overfit to the target image. This diversity can be achieved by altering hairstyle, hair color, skin tone, body pose, or facial expression.

    \item \textbf{Vary the non-target garments to provide outfit diversity.} While the target garment remains unchanged, other garments should be modified. For example, if the target garment is an upper-body item, the reference image should retain the same upper-body garment while altering the lower-body clothing, shoes, or accessories.
\end{itemize}
Since the reference image is edited from the target image $\boldsymbol{p}_{i,\boldsymbol{c}_i}$ in existing datasets, the first requirement ensures faithful garment preservation, while the second and third promote diversity in non-target regions to prevent overfitting and prevent the model from taking the shortcut of directly copying the target image in the generated try-on results.

As illustrated in the pipeline, we use \textit{Flux-kontext}~\cite{labs2025flux1kontextflowmatching} to synthesize reference-person images. Given a target image and carefully designed text prompts, \textit{Flux-kontext} generates images that faithfully preserve the target garment's color, texture, and design while varying the wearer's appearance and non-target clothing. Specifically, we first feed the target image and a textual instruction into \textit{Qwen2.5-VL}~\cite{bai2025qwen2} to obtain a detailed description of the person's appearance, together with an opposite description. We then collect a set of non-target garments and action descriptions to introduce diversity in clothing and pose; examples of these garments and action items are provided in Appendix \ref{appendix:1}. Finally, we concatenate the opposite appearance description with the action and non-target garment descriptions to form the \textit{positive prompt}, while the original appearance description is used as the \textit{negative prompt}. These prompts are then fed into \textit{Flux-kontext} to synthesize the reference images.

We supplement existing virtual try-on benchmarks including VITON-HD~\cite{han2018viton}, DressCode~\cite{morelli2022dress}, ViViD~\cite{fang2024vivid}, FashionTryOn~\cite{zheng2019virtually} and IGPairs~\cite{shen2025imagdressing} by generating corresponding reference pairs for each target garment image $\boldsymbol{c}_i$, forming data pairs $[\boldsymbol{c}_i, \boldsymbol{r}_i]$ that are used for both model training and evaluation.Some open-source datasets, such as FashionTryOn~\cite{zheng2019virtually} and IGPairs~\cite{shen2025imagdressing}, contain numerous duplicated or low-quality samples. We compare the CLIP features of images to filter out redundant samples and employ \textit{Qwen2.5-VL} to identify distorted or unclear images, as well as images where the person faces away from the camera, ensuring overall data quality before generating the final reference set via our data generation pipeline.
Finally, we enrich open-source virtual try-on datasets with additional visual references $\boldsymbol{r}_i$ and person images $\boldsymbol{\bar p}_{i,c_j}$, and combine them to form our own dataset, named \textbf{Virtual Fitting with Reference (VFR)}, for training our RefTon model. The detailed data collection, filtering procedures, and visualization of the samples are provided in Appendix \ref{appendix:1}.

%% file: sec/4_experiment.tex
\begin{figure}[t]
  \centering
  \includegraphics[width=0.98\linewidth]{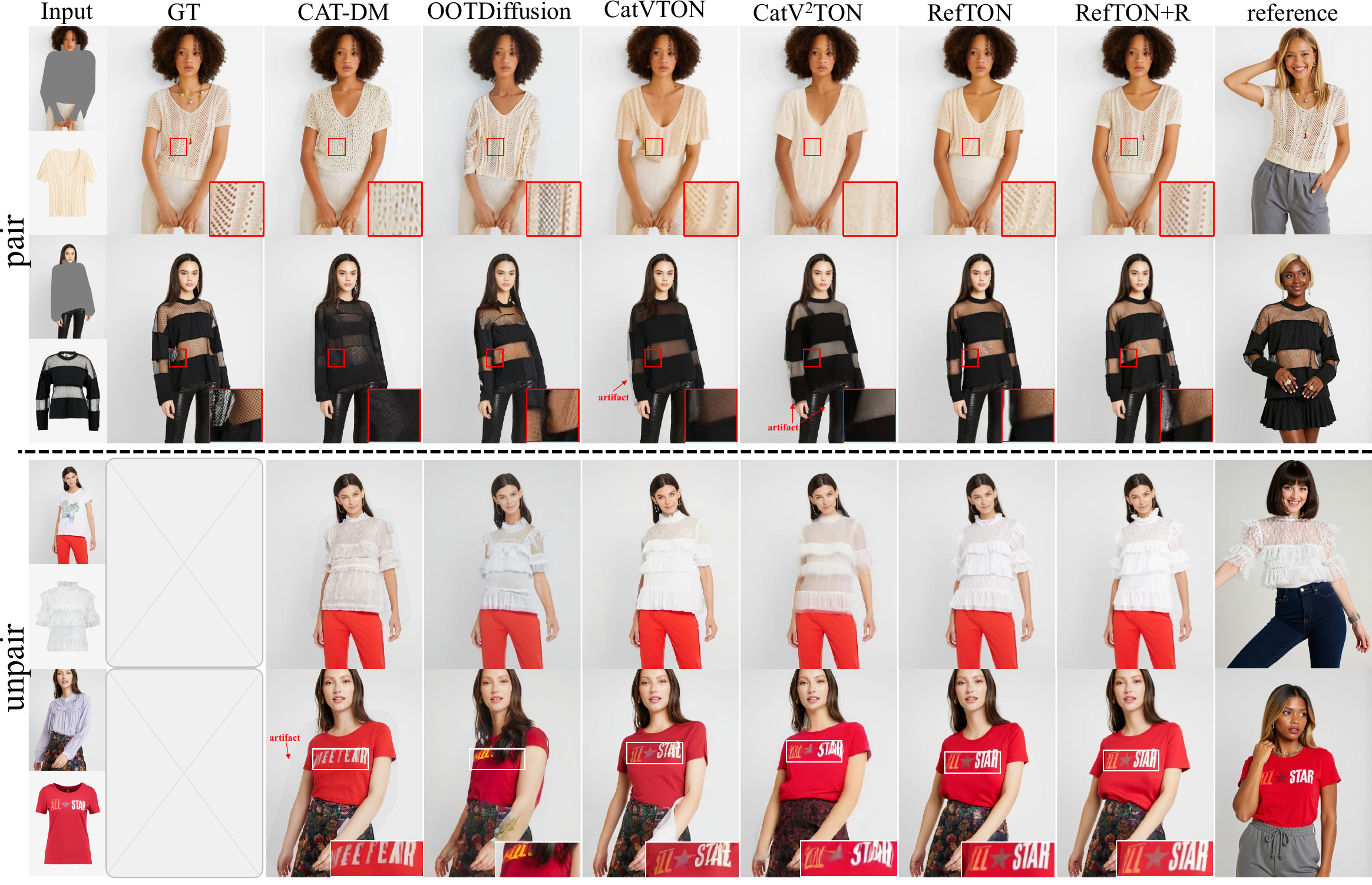} 
  \caption{Qualitative comparison on the VITON-HD dataset. ``reference'' denotes using additional reference $\boldsymbol{r}_i$ for the inference.}
  \label{fig:exp_viton}
\end{figure}

\begin{figure*}[t]
  \centering
  \includegraphics[width=0.98\linewidth]{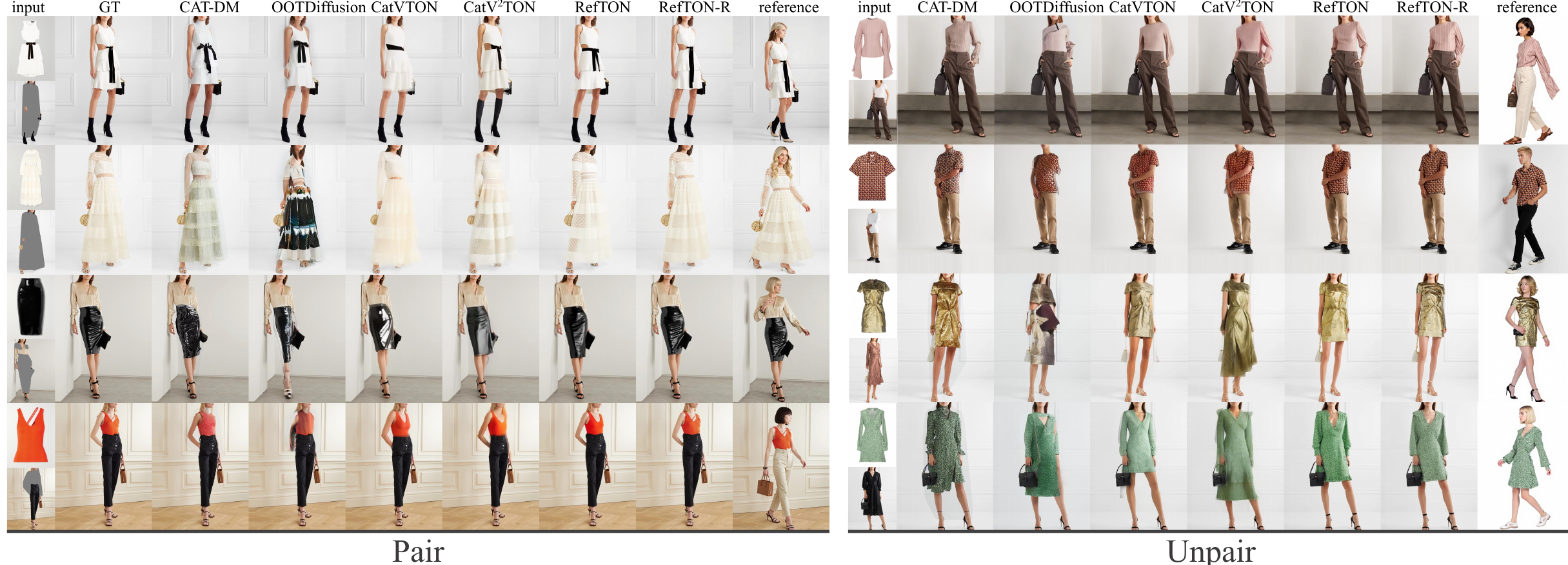} 
\caption{\textbf{Qualitative comparison on the DressCode dataset.}, and the model is trained following the pipeline in Fig.~\ref{Fig:two_stage} (b). ``reference'' denotes the reference $\boldsymbol{r}_i$ image is used during the inference.}
  \label{fig:exp_dresscode}
\end{figure*}
\section{Experiments}
\label{Experiments}
We mainly evaluate our method on two public benchmarks, DressCode~\cite{morelli2022dress} and VITON-HD~\cite{choi2021viton}, both containing images with a resolution of $1024 \times 768$. The VITON-HD comprises 13,670 upper-body image pairs of women, split into 11,647 pairs for training and 2,032 pairs for testing. The DressCode includes three subsets—upper body, lower body, and dresses—with 48,392 training and 5,400 testing pairs. Since DressCode does not provide wrapped cloth masks or agnostic images $\boldsymbol{a}_i$, we generate them using the mask generation tool from CatVTON~\cite{chong2024catvton}. We further conducted experiments on the StreetTryOn~\cite{cui2023street-tryon} and included comparisons with the methods mentioned to demonstrate the model’s P2P and in-the-wild capabilities. In the P2P evaluation, we did not use the cloth-only image as input and only used the person image as a reference. 

During training, we fine-tune our model on the \textit{flux-kontext} backbone using the Low-Rank Adaptation (LoRA) technique, with a rank of $64$ and $\alpha=128$, optimized by \textit{AdamW}. For quantitative evaluation, all generated images are resized to $512 \times 384$ to ensure a fair comparison with previous methods. In a single-dataset experiment, the model is trained independently for 20,000 steps on VITON-HD and 48,000 on DressCode with a batch size of 128, using 8 NVIDIA H100 GPUs. To further enhance generalization and robustness to in-the-wild inputs, we train an additional model on the mixed VFR dataset introduced in Sec.~\ref{section:dataset}. We report cross-dataset evaluation results of our RefTon model at a resolution of $1024 \times 768$ on both VITON-HD and DressCode, and demonstrate its in-the-wild performance using images collected by ourselves.

\subsection{Quantative result}
We provide the numerical results of our model on the VITON-HD and DressCode datasets, distinguishing between paired and unpaired try-on settings. For paired try-on settings with ground truth in test datasets, we utilize four metrics to assess the similarity between the synthesized images and their corresponding authentic images: the Structural Similarity Index \textit{(SSIM)} \cite{nilsson2020understanding}, Learned Perceptual Image Patch Similarity \textit{(LPIPS)} \cite{zhang2018unreasonable}, Fr\'echet Inception Distance \textit{(FID)}\cite{Seitzer2020FID}, and Kernel Inception Distance \textit{(KID)}\cite{binkowski2018demystifying}. For unpaired settings, where we measure the distributional similarity between the synthesized and real samples, we specifically rely on the \textit{Fr\'echet Inception Distance (FID)} and \textit{Kernel Inception Distance (KID)}.  As shown in Table~\ref{tab:combined_reftron}.

\input{tables/table-merge-vtion-dresscode}
Our method (\textbf{RefTon}) consistently performs better than prior baselines, demonstrating higher try-on fidelity and strong alignment with the target person’s pose. With the addition of reference images (``+\textbf{R}''), the quality and detail consistency of the try-on results are further improved compared with the results without reference images, establishing new state-of-the-art results across multiple metrics. Notably, even in the mask-free setting---without agnostic masks or auxiliary inputs---our method maintains garment style correctness and pose consistency, while reaching accuracy on par with or superior to baseline methods, highlighting its robustness and practicality.

Table~\ref{tab:combined_reftron} summarizes the quantitative evaluation on the DressCode dataset. Our method (RefTon) outperforms all baselines, delivering higher try-on quality and achieving strong consistency with the target person’s pose and body structure. Integrating reference images (“+R”) further enhances the results, establishing a new state-of-the-art. Importantly, even in the mask-free setting—without agnostic masks or additional inputs—our method correctly preserves garment styles (e.g., clothing length and design) and maintains high pose alignment, while achieving accuracy comparable to or surpassing prior baselines, demonstrating robustness and practicality. Additional quantitative results on the DressCode subset are provided in Appendix \ref{appendix:2} in detail.

Results in Table~\ref{tab:street-tryon} show that our method achieves state-of-the-art performance on the StreetTryOn benchmark. Notably, our model is neither trained on the StreetTryOn domain nor designed to handle the absence of garment images during training. These results demonstrate its strong generalization ability and flexibility, and further confirm that it can directly extract garment information from the reference image for virtual try-on.
\input{tables/table_street_tyron}

\subsection{Qualitative comparison}

Previous methods can fit garments onto a person, but often fail to preserve wearing-critical attributes such as cut, style, and fine details. By contrast, our method achieves improved visual fidelity among all baselines. As shown in Fig.~\ref{fig:exp_viton}, on VITON, RefTon realistically renders challenging materials such as hollow and semi-transparent fabrics even without reference images. For example, it faithfully preserves the perforated structures and transparency of lace garments, while baselines often produce solid textures or spurious dotted artifacts. Our method also better maintains garment patterns, keeping printed letters and logos clear and consistent with the input. Adding reference images further improves generation quality, validating the effectiveness of our reference assist framework. Even without agnostic masks, directly transferring clothing $\boldsymbol{c}_i$ onto person images $\boldsymbol{\bar p}_{i,\boldsymbol{c}_j}$ still allows our method to outperform most baselines. More results on VITON-HD, DressCode, and in-the-wild examples are provided in Appendix \ref{appendix:3}.

\input{tables/table_abligation}

Our model also performs well on a benchmark with a wider variety of clothing types. As shown in Fig.~\ref{fig:exp_dresscode}, we evaluate our method on the DressCode dataset, where garments are categorized into \textit{upper body}, \textit{lower body}, and \textit{dresses}. Our approach produces more faithful and natural try-on results compared to previous methods. In particular, it renders reflective materials, such as leather and metallic fabrics, with superior realism, avoiding the over-smoothing or distortion artifacts commonly observed in other methods. Moreover, even without agnostic masks, our model can still perform consistent try-on guided by garment style, accurately preserving the length, structure, and overall design without introducing mismatched or inconsistent shapes.

\subsection{Training on VRF Dataset and Evaluation}
To evaluate the effectiveness of our person-to-person virtual try-on framework and data construction pipeline, we build a mixed dataset, \textit{Mixed-Virtual-Ref}, by combining samples from DressCode, VITON-HD, FashionTryOn~\cite{xiao2025omnigen}, ViViD~\cite{fang2024vivid}, and IGPairs~\cite{choi2024improving}. We use \textit{Qwen2.5-VL} to filter low-quality images, generate agnostic masks via~\cite{chong2024catvton}, and obtain reference and unpaired person images through our data generation process. In total, 103{,}936 image pairs are collected for training, using the same hyperparameters as in the DressCode and VITON-HD experiments. 

We evaluate the model on the DressCode and VITON-HD test sets, with quantitative results shown in Table~\ref{tab:combined_dataset}. Our method (RefTon) achieves consistently superior or comparable performance to OOTDiffusion~\cite{xu2025ootdiffusion} across both benchmarks and both paired and unpaired settings. Notably, despite not being trained on DressCode or VITON-HD individually, the mixed-dataset model surpasses dataset-specific baselines on most metrics (e.g., FID and LPIPS), demonstrating strong cross-dataset generalization and the robustness of our data construction strategy.

\subsection{Ablation Study}
\begin{figure}[t]
  \centering
  \includegraphics[width=0.98\linewidth]{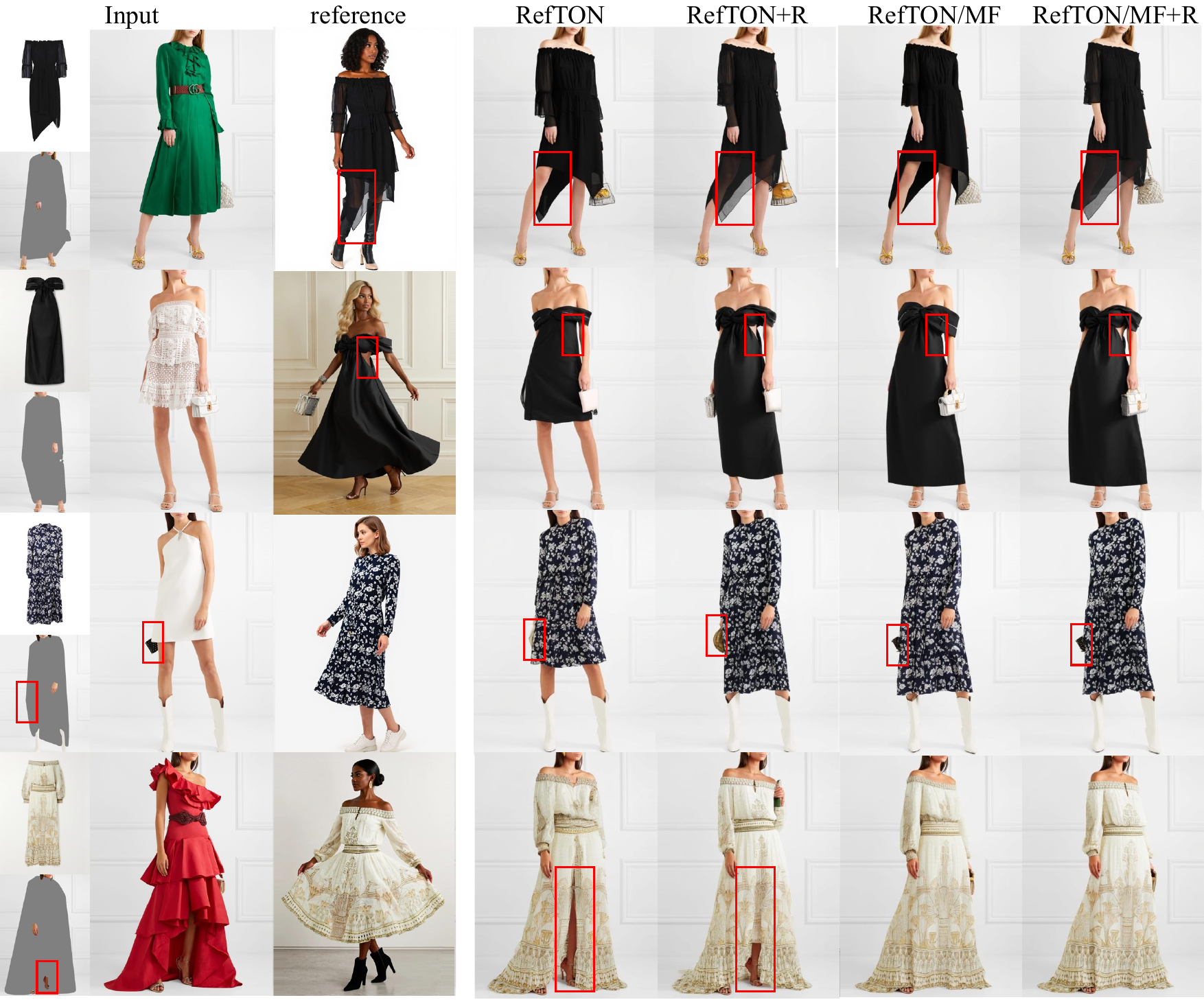} 
\caption{\textbf{Qualitative results of the ablation study across different settings.}
``Ref.'' denotes that a reference image is provided, while ``MF'' indicates mask-free inputs using the original person image instead of a masked agnostic image.}
  \label{fig:ablation}
\end{figure}
We conduct an ablation study to examine our model under four settings(w/\&w/o mask,w/\&w/o Ref.). As shown in Table~\ref{tab:combined_reftron}, our model maintains consistently strong performance across all settings. Introducing a reference image yields clear improvements in both mask-based and mask-free modes, while moving from mask-based to mask-free inputs causes only mild metric fluctuations, confirming the model’s stable robustness without masks.

Fig.~\ref{fig:ablation} further provides qualitative comparisons. The first two rows show that reference images help the model correctly infer garment structures and materials that are ambiguous in the flat images (e.g., hollow textures, semi-transparency). The last two rows illustrate that mask quality heavily affects mask-dependent models: overly aggressive masks remove important items (e.g., handbags), while conservative masks retain unwanted regions (e.g., legs), leading to incorrect garment geometry. In contrast, our mask-free model consistently produces correct outputs regardless of mask or reference conditions, demonstrating that mask-free capability reduces reliance on mask quality and enables more flexible and stable try-on performance.

To validate the rescaled position index, we experiment on VITON-HD with cloth image $\boldsymbol{c}_i$, dense pose $\boldsymbol{d}_i$, warp mask $\boldsymbol{m}_i$, and reference image $\boldsymbol{r}_i$ as conditions. In the masked and mask-free settings, the target cloth is transferred to the person image and the agnostic image, respectively. We evaluate two condition scales, $0.25\times$ and $0.5\times$, where the person or agnostic image is resized to the target resolution and the conditional inputs are rescaled accordingly. As shown in Table~\ref{tab:pos_index}, the rescaled position index outperforms the original \textit{Flux-Kontext} position index on FID and KID under both settings, showing its effectiveness for multi-condition inputs with varying resolutions.

\input{tables/ablation_indx}

\section{Conclusion}
This paper introduces \textbf{RefTon}, a virtual try-on framework that supports both mask-based and mask-free inference and leverages reference images to guide the try-on process. We extend \textit{Flux-Kontext} to handle multi-condition inputs of varying resolutions via a rescaled position index. To train RefTon, we propose a reference data generation pipeline integrating \textit{Qwen2.5-VL} and \textit{Flux-Kontext}. This design allows RefTon to faithfully preserve translucent fabrics, intricate designs, and fine details, consistently outperforming existing methods both quantitatively and qualitatively, achieving state-of-the-art performance in all settings.


%% file: tables/table-merge-vtion-dresscode.tex
\begin{table*}[htbp]
\caption{
Quantitative comparison on VITON-HD~\cite{choi2021viton} and DressCode~\cite{morelli2022dress}. 
Best and second-best results are shown in \textbf{bold} and \underline{underline}, respectively. 
``+R'' denotes the use of reference images, ``MF'' indicates mask-free inputs, and ``--'' denotes missing values. 
Subscripts $p$ and $u$ denote the \textit{paired} and \textit{unpaired} settings, respectively.
}
\label{tab:combined_reftron}
\centering
\resizebox{\textwidth}{!}{
\begin{tabular}{lcc|cccccc||cccccc}
\toprule
\multirow{3}{*}{\textbf{Method}} 
& \multicolumn{2}{c|}{\textbf{Input}} 
& \multicolumn{6}{c||}{\textbf{VITON-HD}} 
& \multicolumn{6}{c}{\textbf{DressCode}} 
\\
\cmidrule(lr){2-3} \cmidrule(lr){4-9} \cmidrule(lr){10-15}
& \textbf{Mask} & \textbf{Pose}
& \textbf{LPIPS}$_p\downarrow$
& \textbf{SSIM}$_p\uparrow$
& \textbf{FID}$_p\downarrow$
& \textbf{KID}$_p\downarrow$
& \textbf{FID}$_u\downarrow$
& \textbf{KID}$_u\downarrow$
& \textbf{LPIPS}$_p\downarrow$
& \textbf{SSIM}$_p\uparrow$
& \textbf{FID}$_p\downarrow$
& \textbf{KID}$_p\downarrow$
& \textbf{FID}$_u\downarrow$
& \textbf{KID}$_u\downarrow$
\\
\midrule
CAT-DM\(^*\)~\cite{zeng2024cat}
& \checkmark & \checkmark
& 0.080 & 0.877 & 5.60 & 0.83 & 8.93 & 1.37
& -- & -- & -- & -- & -- & --
\\

IDM-VTON\(^*\)~\cite{choi2024improving}
& \checkmark & \checkmark
& 0.102 & 0.870 & 6.29 & -- & -- & --
& 0.062 & 0.920  & 8.64  & 0.90  & --  & --
\\

OOTDiffusion\(^*\)~\cite{xu2025ootdiffusion}
& \checkmark & --
& 0.071 & 0.878 & 8.81 & 0.82 & -- & --
& 0.045 & \textbf{0.927} & 4.20 & \textbf{0.37} & -- & --
\\

CatVTON\(^*\)~\cite{chong2024catvton}
& \checkmark & --
& \underline{0.057} & 0.870 & 5.43 & 0.41 & 9.02 & 1.09
& 0.046 & 0.892 & 3.99 & \underline{0.82} & 6.14 & 1.40
\\

CatVT2ON\(^*\)~\cite{chong2025catv2ton}
& \checkmark & \checkmark
& \underline{0.057} & \textbf{0.890} & 8.10 & 2.25 & 11.22 & 2.99
& \underline{0.037} & \underline{0.922} & 5.72 & 2.34 & 8.63 & 3.84
\\

OmniVTON\(^*\)~\cite{yang2025omnivton}
& \checkmark & \checkmark
& 0.145 & 0.832 & 7.76 & -- & 9.62 & --
& 0.119 & 0.865 & 5.34 & -- & 6.45 & --
\\

PromptDresser$_{pose}^*$~\cite{kim2024promptdresser}
& \checkmark & \checkmark
& 0.097 & \underline{0.878} & 9.07 & 1.16 & -- & --
& -- & -- & -- & -- & -- & --
\\

PromptDresser\(^*\)~\cite{kim2024promptdresser}
& \checkmark & --
& 0.112 & 0.869 & 8.54 & \textbf{0.67} & -- & --
& -- & -- & -- & -- & -- & --
\\

\rowcolor{lightgreen}
\textbf{RefTON (Ours)}
& \checkmark & --
& \underline{0.057} & 0.873 & \underline{5.45} & 0.82 & \underline{8.58} & \underline{1.06}
& \underline{0.037} & 0.912 & \underline{3.48} & 1.20 & \underline{5.31} & \underline{1.36}
\\

\rowcolor{lightgreen}
\textbf{RefTON+R (Ours)}
& \checkmark & --
& \textbf{0.049} & \underline{0.879} & \textbf{4.69} & \underline{0.68} & \textbf{8.43} & \textbf{0.91}
& \textbf{0.031} & 0.918 & \textbf{2.94} & 0.95 & \textbf{5.07} & \textbf{1.15}
\\
\midrule
\multicolumn{15}{c}{\textit{Mask-Free setting}} \\
\midrule 
CatVTON(Mask-Free)\(^*\)~\cite{chong2024catvton}
& -- & --
& \underline{0.061} & \underline{0.870} & \underline{5.89} & \textbf{0.51} & 9.29 & 1.17
& 0.045 & \underline{0.902} & 4.78 & \underline{1.30} & 7.40 & 2.62 \\

Any2AnyTryon\(^*\)~\cite{guo2025any2anytryon}
& -- & --
& 0.088 &  0.839 & 6.93 & \underline{0.74} & 8.97 & 0.98
& -- & -- & -- & -- & -- & -- \\

TryOffDiff\(^*\)~\cite{velioglu2025enhancing} 
& -- & --
& -- & -- & -- & -- & 11.9 & 2.60 
& -- & -- & -- & -- & 7.90 & 2.70
\\

\rowcolor{lightgreen}
   \textbf{RefTON/MF} (Ours)
   & -- & --
      & \underline{0.061}  & 0.866  & 5.98  & 1.04  & \underline{8.40} & \underline{0.81}
     & \underline{0.041}  & 0.901   & \underline{3.84}   & 1.33   & \textbf{5.00}  & \textbf{1.17}  \\
      \rowcolor{lightgreen}
   \textbf{RefTON+R/MF}(Ours)
    & -- & --
      & \textbf{0.053}  & \textbf{0.872}  & \textbf{5.11}  & \underline{0.82}  & \textbf{8.32} & \textbf{0.78} 
     & \textbf{0.035}  & \textbf{0.906}   & \textbf{3.34}   & \textbf{1.15}   & \underline{5.02}  & \underline{1.28}  \\
\bottomrule
\end{tabular}}
\end{table*}

%% file: tables/table_street_tyron.tex
\renewcommand{\arraystretch}{0.9}
\begin{table}[t]
\caption{Quantitative comparisons on the StreetTryOn. Sh, St, P, denotes the shop, street, and person(model) image, respectively.}
\label{tab:street-tryon}
\centering
\resizebox{0.99\linewidth}{!}{%
\begin{tabular}{lccc||cccc}
\toprule
\multirow{2}{*}{\textbf{Method}} 
& \multicolumn{3}{c||}{\textbf{Required Input}} 
& \textbf{Sh-to-St} 
& \textbf{P-to-P} 
& \textbf{P-to-St}
& \textbf{St-to-St} \\
\cmidrule(lr){2-4}\cmidrule(lr){5-8}
& \textbf{Mask} & \textbf{Pose} & \textbf{Text}
& \textbf{FID}$\downarrow$
& \textbf{FID}$\downarrow$
& \textbf{FID}$\downarrow$
& \textbf{FID}$\downarrow$ \\
\midrule
StreetTryOn & \cmark & \cmark & \xmark 
& 34.054 & 12.185 & 34.191 & 33.039 \\
OmniVTON
& \cmark & \cmark & \cmark
& 33.919 & 8.983 & 33.450 & 23.470 \\
\rowcolor{lightgreen}
\textbf{RefTON}
& \cmark & \xmark & \xmark
& \textbf{28.991} & \textbf{8.870} & \textbf{25.429} & \textbf{16.452}\\

\bottomrule
\end{tabular}%
}
\end{table}

%% file: tables/table_abligation.tex
\begin{table}[!h]
    \centering
    \caption{
    Cross-dataset comparison with OOTDiffusion~\cite{xu2025ootdiffusion} on VITON-HD and DressCode under both paired and unpaired settings.
    For the mask-free setting, we report only the unpaired results, as the paired setting is not meaningful when no garment--person alignment is enforced.
    }
    \resizebox{0.47\textwidth}{!}{
    \setlength{\tabcolsep}{1mm}
    \begin{tabular}{l|cccc|cccc}
        \toprule
        \multirow{2}{*}{\textbf{Methods}} 
        & \multicolumn{4}{c|}{\textbf{VITON-HD (Paired)}} 
        & \multicolumn{4}{c}{\textbf{DressCode (Paired)}} \\
        \cmidrule(lr){2-5} \cmidrule(lr){6-9}
        & \textbf{SSIM}$\uparrow$  & \textbf{FID}$_p$$\downarrow$ & \textbf{KID}$_p$$\downarrow$ & \textbf{LPIPS}$\downarrow$
        & \textbf{SSIM}$\uparrow$  & \textbf{FID}$_p$$\downarrow$ & \textbf{KID}$_p$$\downarrow$ & \textbf{LPIPS}$\downarrow$ \\
        \midrule
        OOTDiffusion\(^*\)~\cite{xu2025ootdiffusion}
        & 0.839 & 11.22 & 2.72 & 0.123
        & 0.915 & 11.96 & 1.21 & 0.061 \\
        \midrule
        \rowcolor{lightgreen}
        \textbf{RefTON} (Ours)
        & 0.851 & 6.23 & 0.80 & 0.072
        & 0.896 & 3.70 & 1.13 & 0.045 \\
        \rowcolor{lightgreen}
        \textbf{RefTON+R} (Ours)
        & \textbf{0.859} & \textbf{5.13} & \textbf{0.62} & \textbf{0.060}
        & \textbf{0.903} & \textbf{3.14} & \textbf{0.97} & \textbf{0.038} \\
        \bottomrule
    \end{tabular}
    }

    \vspace{1.0mm}
    \resizebox{0.47\textwidth}{!}{
    \setlength{\tabcolsep}{1mm}
    \begin{tabular}{l|cc|cc}
        \toprule
        \multirow{2}{*}{\textbf{Methods}} 
        & \multicolumn{2}{c|}{\textbf{VITON-HD (Unpaired)}} 
        & \multicolumn{2}{c}{\textbf{DressCode (Unpaired)}} \\
        \cmidrule(lr){2-3} \cmidrule(lr){4-5}
        & \textbf{FID}$_u$$\downarrow$ & \textbf{KID}$_u$$\downarrow$
        & \textbf{FID}$_u$$\downarrow$ & \textbf{KID}$_u$$\downarrow$ \\
        \midrule
        \rowcolor{lightgreen}
        \textbf{RefTON} (Ours)
        & 9.11  & 1.08 
        & 5.22  & 1.20 \\
        \rowcolor{lightgreen}
        \textbf{RefTON+R} (Ours)
        & 8.59  & 0.87 
        & 5.03  & 1.11 \\
        \rowcolor{lightgreen}
        \textbf{RefTON/MF} (Ours)
        & 8.88  & 0.82 
        & 5.03  & 1.23 \\
        \rowcolor{lightgreen}
        \textbf{RefTON+R/MF} (Ours)
        & \textbf{8.39} & \textbf{0.65} 
        & \textbf{4.87} & \textbf{1.10} \\
        \bottomrule
    \end{tabular}
    }
    \label{tab:combined_dataset}
\end{table}

%% file: tables/ablation_indx.tex
\begin{table}[t]
\centering
\caption{
\textbf{Rescaled position index (PI) ablation.}.
We compare the original Flux-Kontext position index with our rescaled PI at different condition scales; $MF$ denotes mask-free inputs. Conditional inputs are resized to $0.5\times$ and $0.25\times$ of the target resolution.
}
\resizebox{0.98\columnwidth}{!}{
\setlength{\tabcolsep}{2mm}
\begin{tabular}{l|cccc|cccc}
\toprule
\multirow{2}{*}{\textbf{Method}} 
& \multicolumn{4}{c|}{\textbf{0.5 Scale}} 
& \multicolumn{4}{c}{\textbf{0.25 Scale}} \\
\cmidrule(lr){2-5} \cmidrule(lr){6-9}
& FID$\downarrow$ & KID$\downarrow$ & FID$_{MF}$$\downarrow$ & KID$_{MF}$$\downarrow$
& FID$\downarrow$ & KID$\downarrow$ & FID$_{MF}$$\downarrow$ & KID$_{MF}$$\downarrow$ \\
\midrule
w/ o Rescaled PI
& 5.29 & 0.84 & 4.75 & 0.72
& 6.08 & 0.93 & \textbf{5.35} & 0.76 \\
\midrule
\rowcolor{lightgreen}
w/ Rescaled PI
& \textbf{5.09} & \textbf{0.79} & \textbf{4.71} & \textbf{0.69}
& \textbf{6.01} & \textbf{0.77} & 5.37 & \textbf{0.71} \\
\bottomrule
\end{tabular}
\label{tab:pos_index}
}
\end{table}

%% file: sec/5_supplentary.tex
\clearpage
\appendix
\renewcommand{\thefigure}{S\arabic{figure}}
\renewcommand{\thetable}{S\arabic{table}}
\setcounter{page}{1}
\maketitlesupplementary
\setcounter{section}{0}  

\section{Generation of Reference Image}
\label{appendix:1}
This section provides a detailed description of the process for generating the reference image. Many virtual try-on datasets offer the garment image and the image of the target person wearing the target garment, but they do not include the reference image, which shows the visual effect of the target garment $\boldsymbol{c}_i$ being worn by another person $\boldsymbol{p}_{\boldsymbol{c}_j}$. The generation of the reference image can be viewed as an image editing task on $\boldsymbol{p}_{\boldsymbol{c}_j}$. As discussed in Section 3.3, the reference image $\boldsymbol{r}_i$ must satisfy three key requirements.

Firstly, the reference image should faithfully preserve the details of the target garment $\boldsymbol{c}_i$. This requires the editing model to have strong consistency abilities. The \textit{Flux-Kontext} model has a robust ability to edit the target region corresponding to the text prompt while keeping unrelated regions—such as the area not related to the garment—unchanged. Moreover, the \textit{Flux-Kontext} model can perform precise local editing according to the text prompt, in this case, focusing on the person under the garment. Therefore, we choose the popular \textit{Flux-Kontext} model [36]
to edit the input image conditioned on the text prompt. Specifically, we add a sentence such as ``keep the \{target cloth\} cloth unchanged'' in the text prompt.

Secondly, the person in the reference image should look significantly different from the person in the target image. During training, we observed that if the reference image is too similar to the target image, the model tends to rely on a "shortcut"—directly copying from the reference image and ignoring the agnostic/person image $\boldsymbol{a}_i$/$\boldsymbol{p}_i$ and the cloth $\boldsymbol{c}_i$. To avoid this, we ensure that the person in the reference image differs from the target image to better showcase the visual effect of the clothing when worn, rather than focusing on the appearance of the person themselves. To achieve this, we utilize the Text-to-Image (T2I) capabilities of the \textit{Flux-Kontext} model. We extract an accurate description of the person's appearance in the target image (e.g., "The model has an East Asian appearance, with light skin, long black hair, and a neutral expression...") and pass it as the \textbf{negative prompt} to the T2I model. In contrast, we provide an opposite description (e.g., "The model has an African appearance, with dark skin, short yellow hair, and a cheerful expression...") as the \textbf{positive prompt} to guide the editing of the target image, as shown in Figure 4(b). 

However, extracting descriptions for each image manually is labor-intensive. To address this, we use a vision-language model (VLM) such as \textit{Qwen2.5-VL} to automatically generate the description and its opposite. Specifically, we pass the target image $\boldsymbol{p}_{i, \boldsymbol{c}_i}$ to the VLM and provide a prompt like ``Start with Positive: describe only the model’s race, skin, hair, eyes, and expression, then give the opposite in one sentence with Negative: changing those traits without 'not' or clothing.'' The generated description and its opposite are then fed into the negative and positive prompt encoders of the T2I model to edit the target image, as shown in Figure 4(a). In this way, we automate the generation of the reference image $\boldsymbol{r}_i$ by editing the target image.

Thirdly, after extracting the description and opposite description of the person's appearance, we introduce more diversity into the reference image by varying the non-target garments and actions of the human in the target image. This can be achieved through the image editing model by adding descriptions related to non-target garments and actions. We provide a description bank containing candidate descriptions for outfits and actions across three scenarios: the person in the image is wearing a dress, an upper-body garment, or a lower-body garment. This ensures that the description of the editable garment differs from the target garment $\boldsymbol{c}_i$. Furthermore, the outfit descriptions also include accessories such as glasses, wristwatches, and bracelets to increase diversity. These descriptions, along with the actions and outfit details, are concatenated into positive prompts and passed to the T5 text encoder, as shown in Figure 4.

\begin{figure*}[ht]
  \centering
  \includegraphics[width=0.99\linewidth, height=0.55\linewidth]{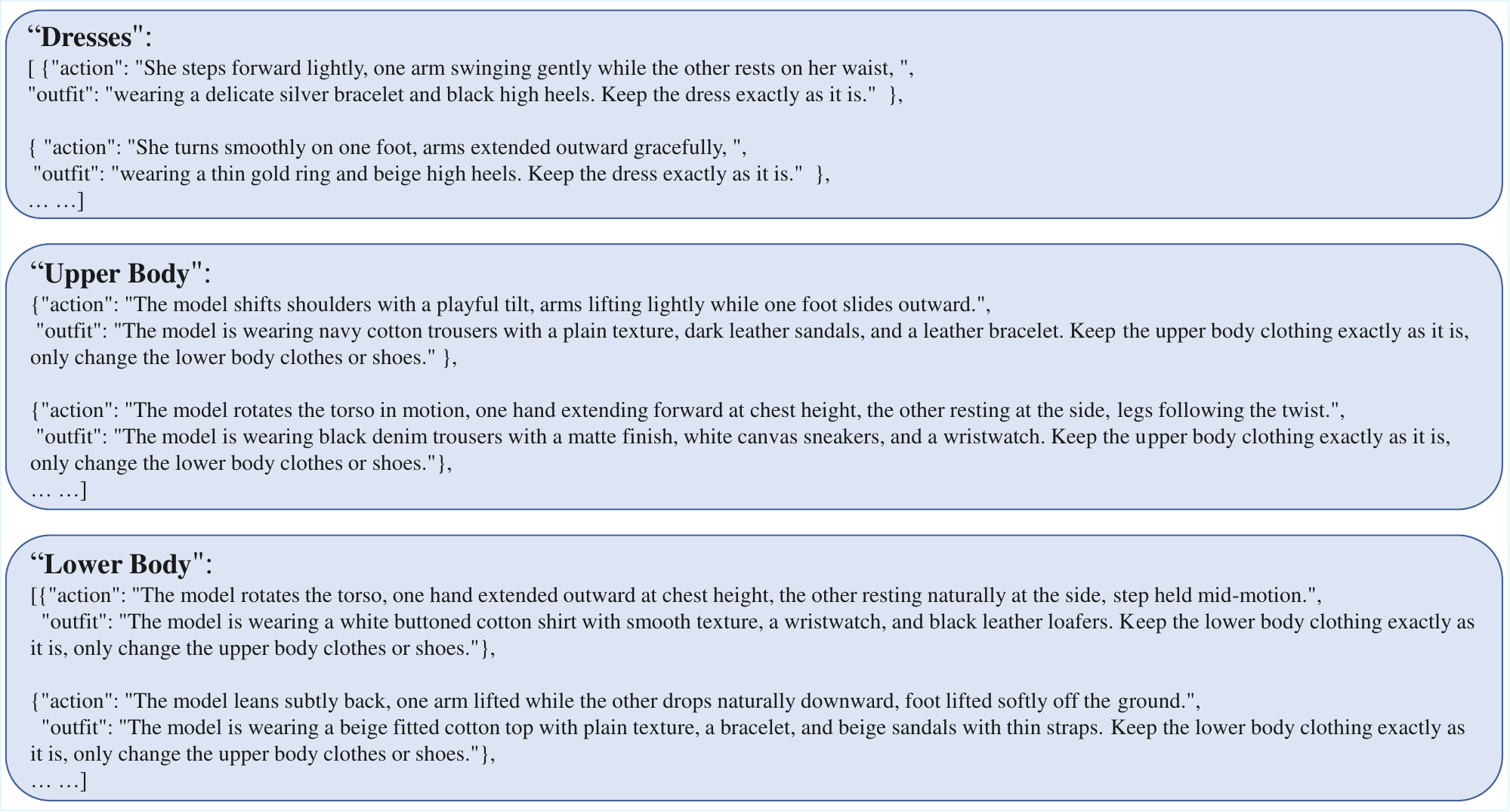} 
\caption{Sample text prompts from the Outfit and Action Description Bank. To ensure the model edits only the person while preserving the target clothing, we assign different outfits and action description categories to different clothing inputs.}
  \label{fig:prompt_bank}
\end{figure*}

\begin{figure*}[ht]
  \centering
  \includegraphics[width=0.98\linewidth, height=0.55\linewidth]{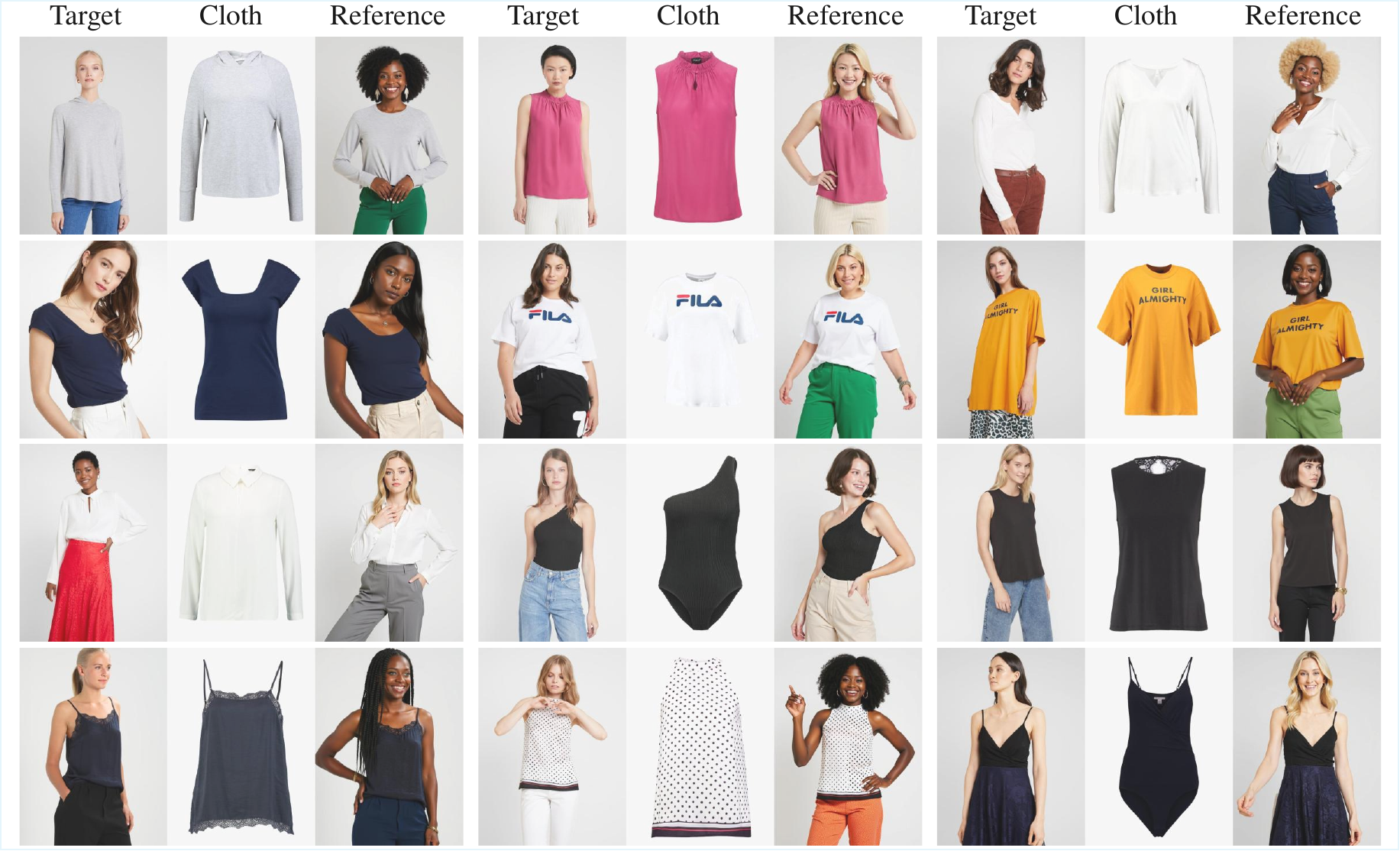} 
\caption{Sample reference images generated by our reference data generation pipeline. The editing model takes the target person's image as input and synthesizes corresponding reference images, while preserving the garment's appearance to match the cloth.}
  \label{fig:reference images}
\end{figure*}

Fig~\ref{fig:prompt_bank} illustrates selected text prompts from the prompt description bank used for reference image generation. Furthermore, Fig~\ref{fig:reference images} exhibits a selection of the resulting reference data samples.

\section{Further Quantitative Evaluation on High Resolution}
\label{appendix:2}

Here we provide more results of the generated image. Table~\ref{tab:dresscode-detail} summarizes the detailed quantitative evaluation on the DressCode dataset. Our method (RefTON) outperforms all baselines, delivering higher try-on quality and achieving strong consistency with the target person’s pose and body structure. Integrating reference images (“+R”) further enhances the results, establishing a new state-of-the-art. Importantly, even in the mask-free setting—without agnostic masks or additional inputs—our method correctly preserves garment styles (e.g., clothing length and design) and maintains high pose alignment, while achieving accuracy comparable to or surpassing prior baselines, demonstrating robustness and practicality.
\input{tables/table_dresscode_detail}

\input{tables/table_merge_appendix}
To further evaluate our method under high-resolution settings, Table~\ref{tab:combined_reftron_appendix} reports quantitative results on both VITON-HD and DressCode at a resolution of 1024. Across the paired and unpaired protocols, RefTON and RefTON+R consistently achieve high performance on nearly all metrics. Notably, the mask-free variants (RefTON/MF and RefTON+R/MF) deliver particularly strong results. These results demonstrate that our framework scales effectively to high-resolution synthesis and remains robust across diverse virtual try-on settings.

\section{Additional Qualitative Results}
\label{appendix:3}
In this section, we provide extensive qualitative visualizations to further demonstrate the robustness, generalization ability, and high-fidelity performance of our method across different datasets, clothing categories, and evaluation settings.

\begin{figure*}[h]
\centering
  \includegraphics[width=0.92\textwidth, height=0.74\textwidth]{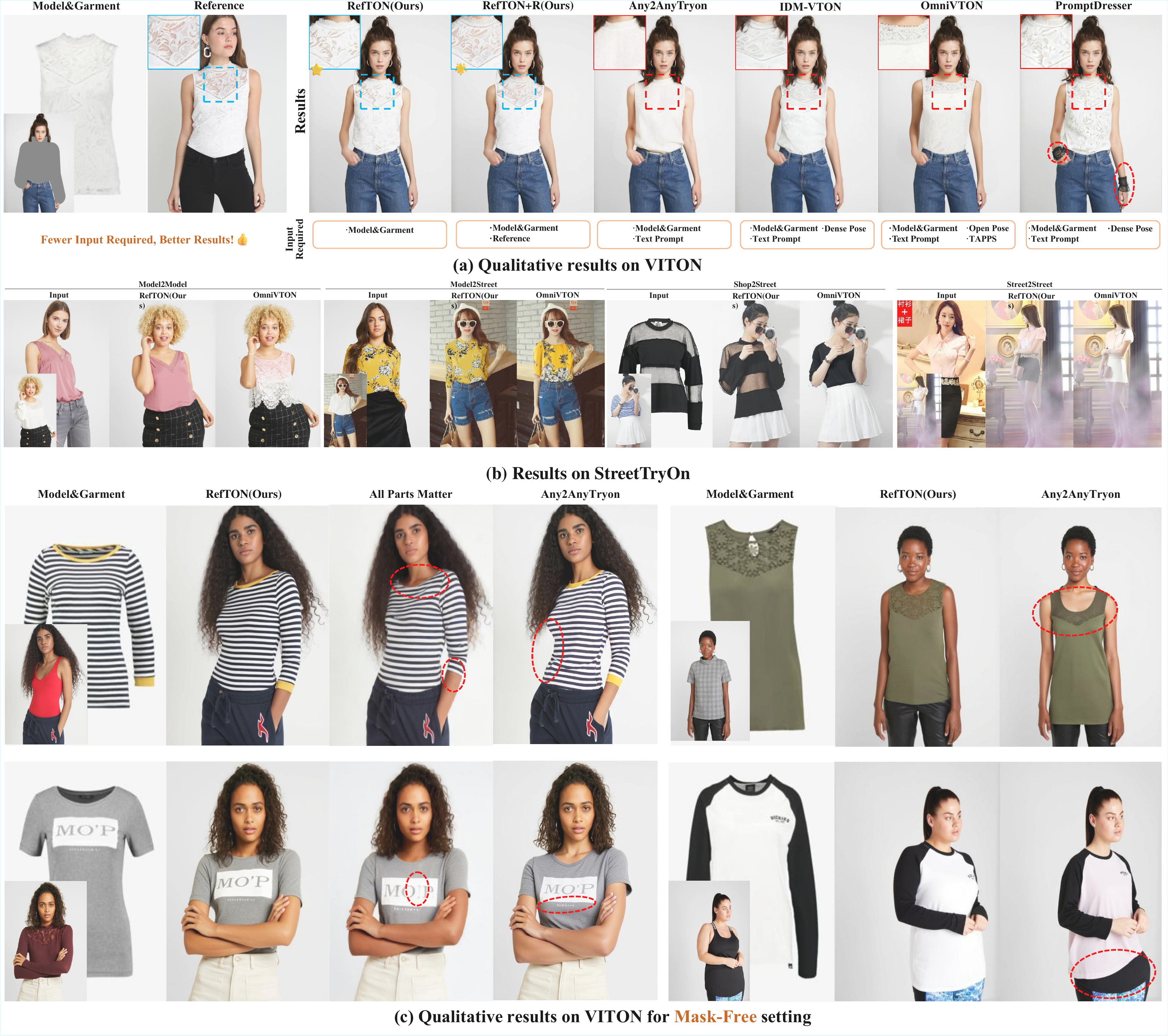}
  \captionof{figure}{(a)We gain better results with fewer inputs required on VITON. (b)StreetTryOn qualitative results (in-the-wild \& P2P): our method yields better results.(c) Our model shows stronger mask-free capability.}
  \label{fig:onecol}
\end{figure*}

We present qualitative comparisons with methods that require additional text inputs, and explicitly annotate each method’s required inputs in Fig~\ref{fig:onecol}(a). Our method achieves the best visual quality with the fewest required inputs. Moreover, adding a reference image further improves intricate details (e.g., lace, transparency, and texture), highlighting the advantage of visual reference. 
Figure~\ref{fig:onecol}(b) shows qualitative results on the StreetTryOn dataset under various settings. Compared with the baselines, our model produces more faithful try-on details while better preserving the person’s pose and the background. We also provide qualitative comparisons under the mask-free setting, as shown in Fig.~\ref{fig:onecol}(c). Our model demonstrates the strongest mask-free virtual try-on capability among all compared methods.

\begin{figure*}[ht]
  \centering
  \includegraphics[width=0.85\linewidth, height=1.25\linewidth]{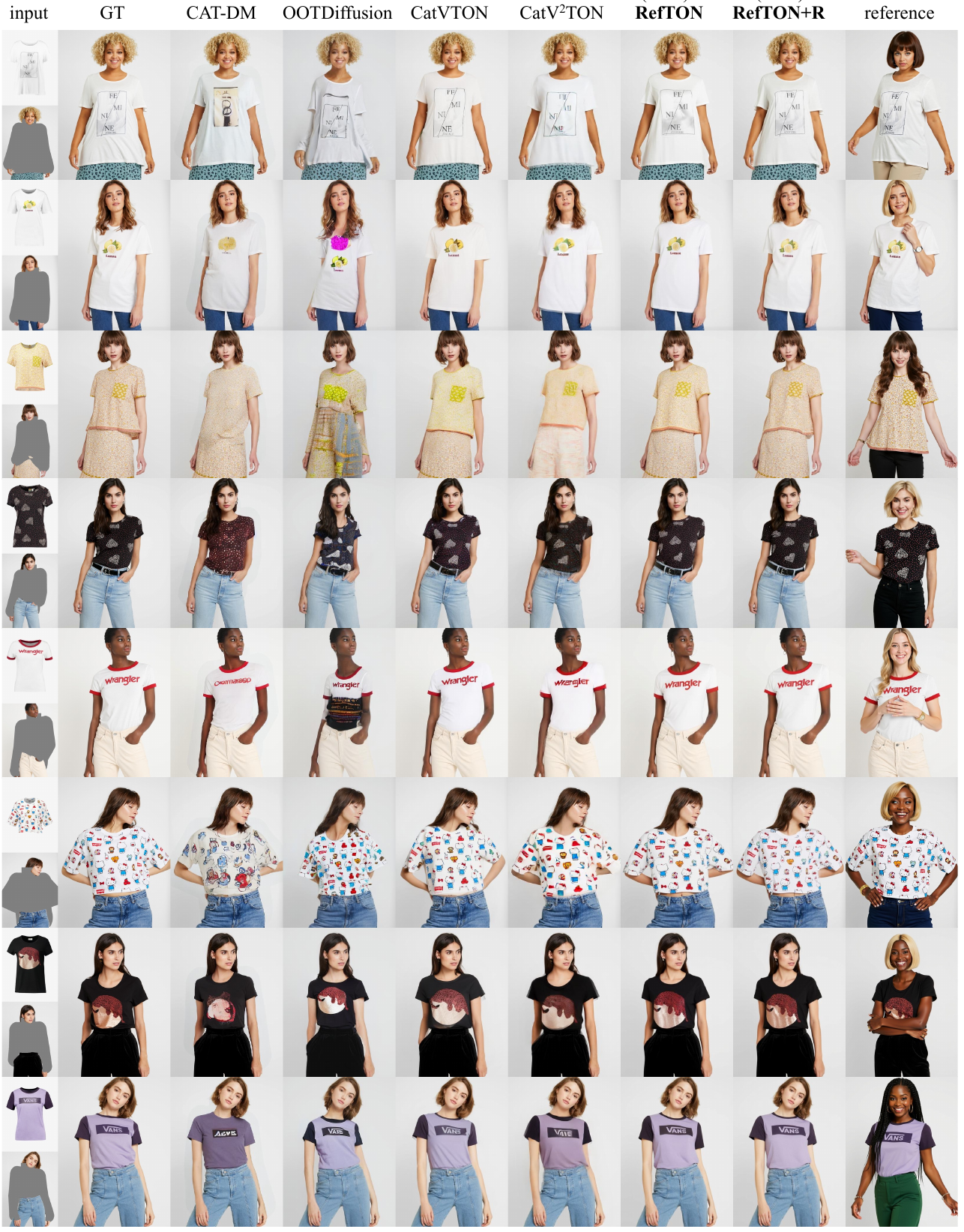} 
\caption{\textbf{Qualitative paired results in VITON-HD dataset with complex patterns on clothes.}
``reference'' denotes that a reference image is provided.}
  \label{fig:supp-vis-viton-patten}

\end{figure*}

\begin{figure*}[ht]
  \centering
  \includegraphics[width=0.85\linewidth, height=1.25\linewidth]{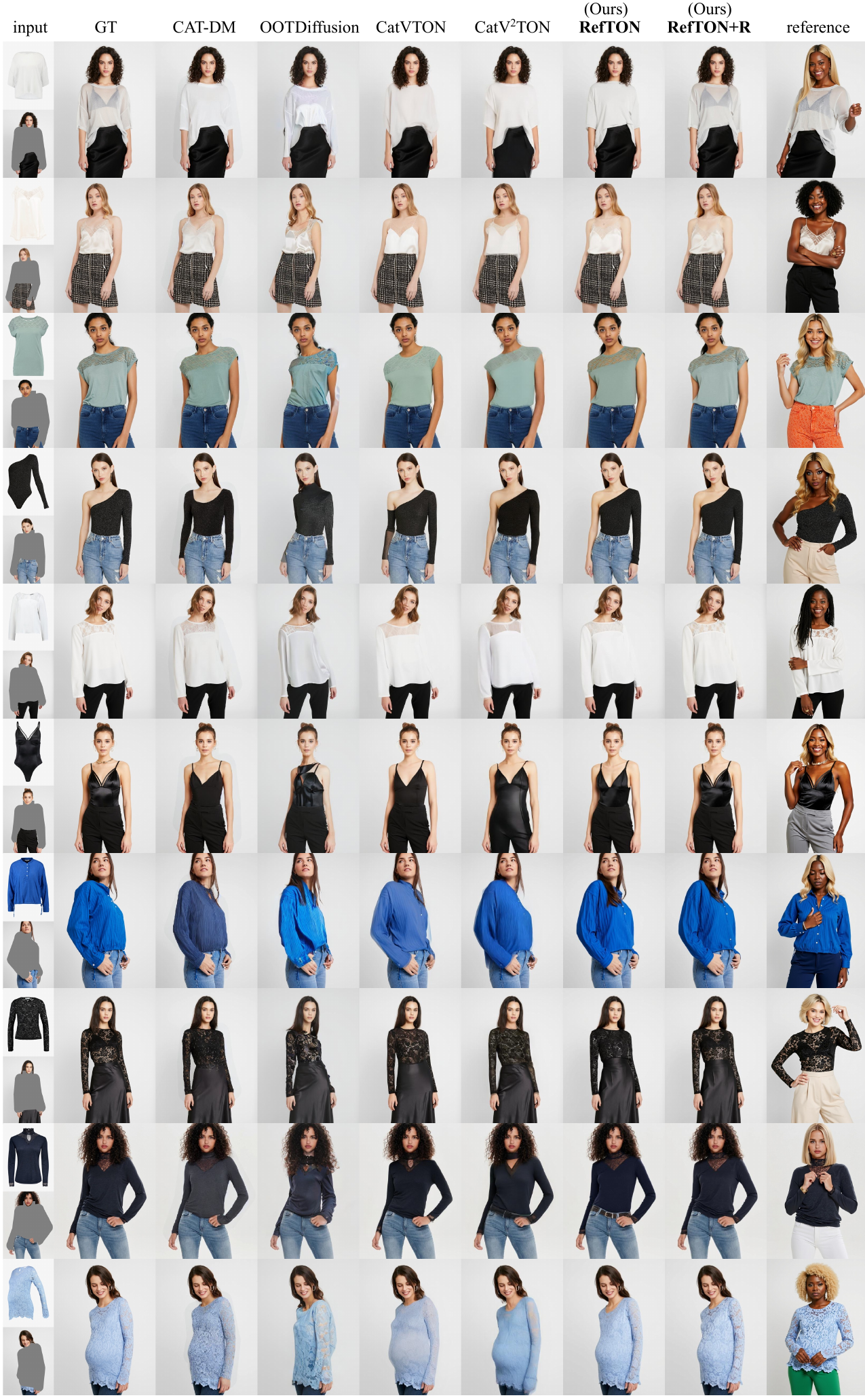} 
\caption{\textbf{Qualitative paired results in VITON-HD dataset with complex structure on clothes.}
``reference'' denotes that a reference image is provided.}
  \label{fig:supp-vis-viton-structure}

\end{figure*}

\begin{figure*}[ht]
  \centering
  \includegraphics[width=0.9\linewidth, height=1.25\linewidth]{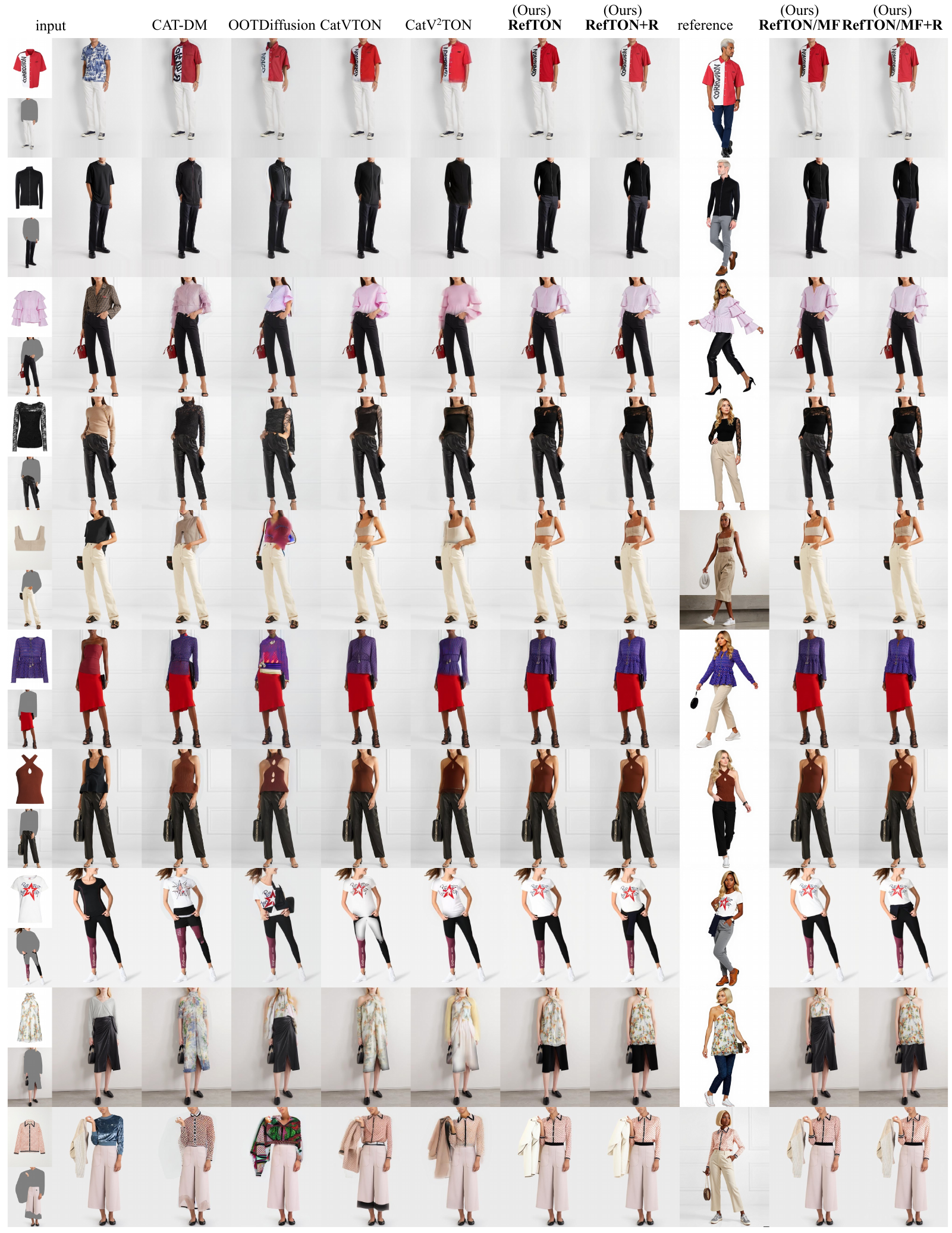} 
\caption{\textbf{Qualitative results of upper-body sub-set in Dresscode dataset unpaired setting.}
``reference'' denotes that a reference image is provided, while ``MF'' indicates mask-free inputs using the original person image instead of a masked agnostic image.}
  \label{fig:supp-vis-upper_body}

\end{figure*}

\begin{figure*}[ht]
  \centering
  \includegraphics[width=1.0\linewidth]{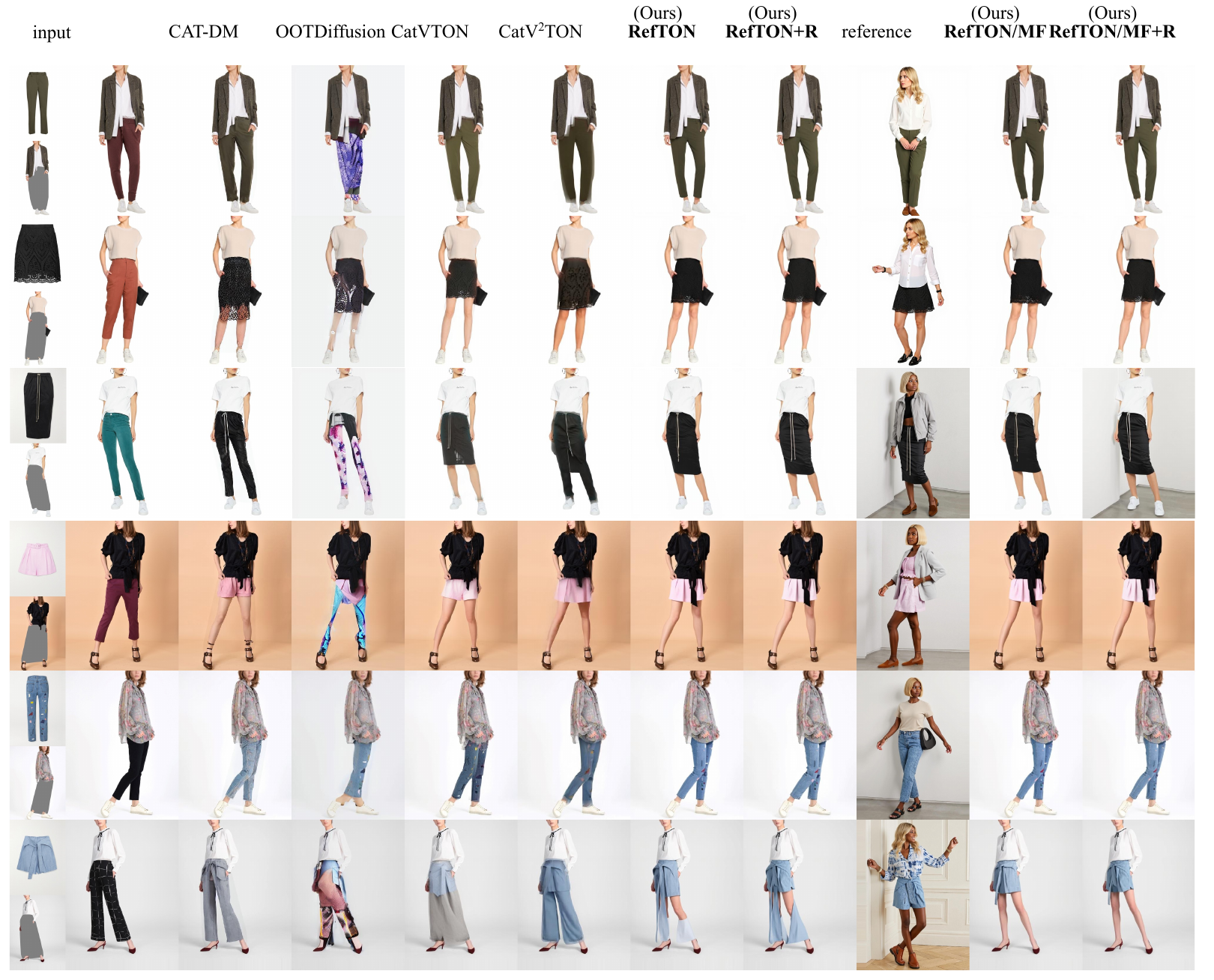} 
\caption{\textbf{Qualitative results of lower-body sub-set in Dresscode dataset unpaired setting.}
``reference'' denotes that a reference image is provided, while ``MF'' indicates mask-free inputs using the original person image instead of a masked agnostic image.}
  \label{fig:supp-vis-lower_body}
\end{figure*}

\begin{figure*}[ht]
  \centering
  \includegraphics[width=1.0\linewidth]{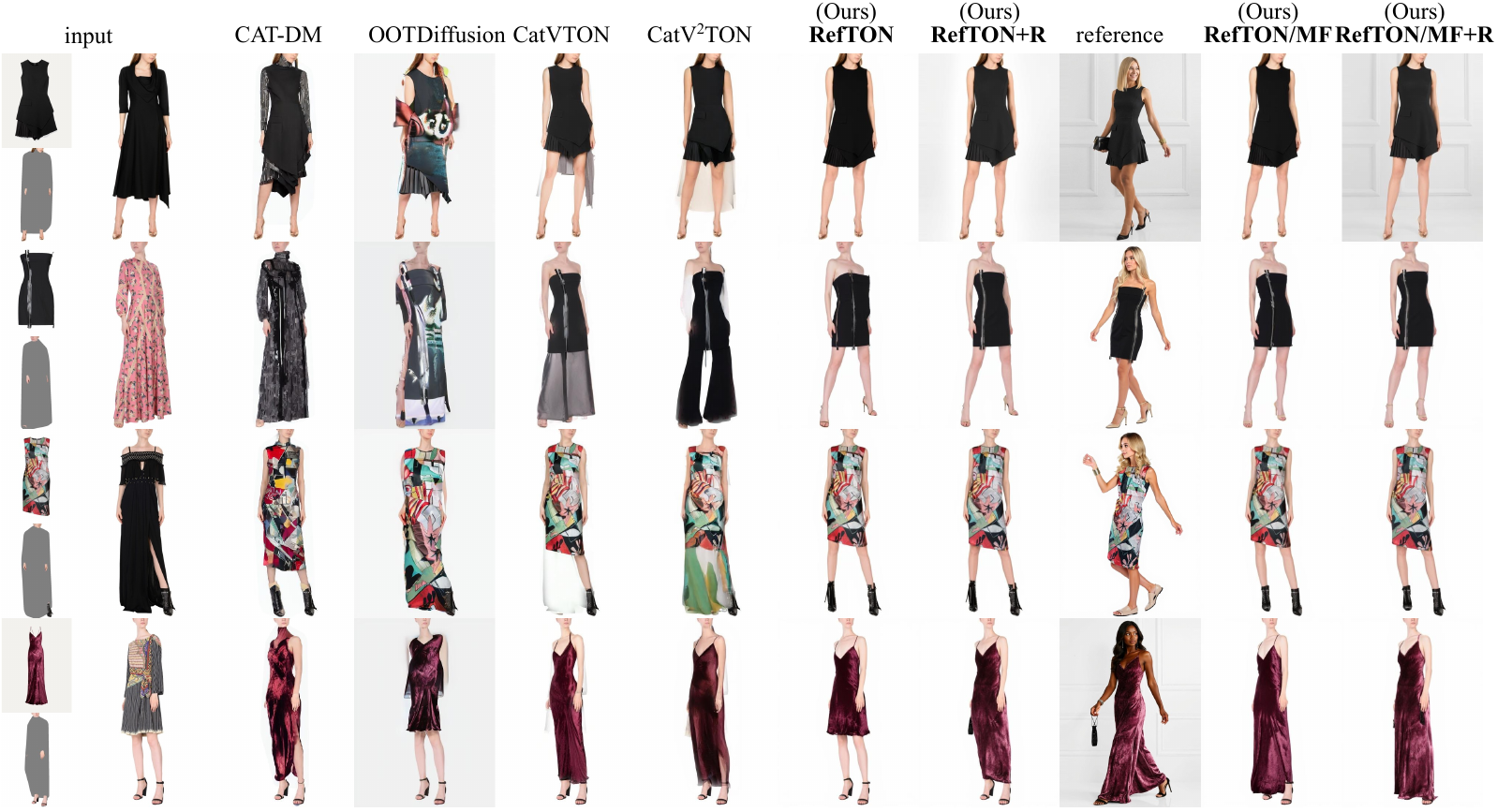} 
\caption{\textbf{Qualitative results of dresses sub-set in Dresscode dataset unpaired setting.}
``reference'' denotes that a reference image is provided, while ``MF'' indicates mask-free inputs using the original person image instead of a masked agnostic image.}
  \label{fig:supp-vis-dress}

\end{figure*}

As shown in Fig.~\ref{fig:supp-vis-viton-patten}, \ref{fig:supp-vis-viton-structure}, \ref{fig:supp-vis-upper_body}, \ref{fig:supp-vis-lower_body}, and \ref{fig:supp-vis-dress}, our approach consistently preserves garment details, structural correctness, and texture realism under both paired and unpaired scenarios, with or without mask-free (MF) inputs. Moreover, as illustrated in Fig.\ref{fig:in-the-wild}, even in challenging in-the-wild conditions, our model exhibits strong robustness—maintaining accurate body pose, preserving background integrity, and producing stable try-on results without introducing artifacts or unintended changes.

\paragraph{Complex Patterns (VITON-HD).}
As shown in Fig.~\ref{fig:supp-vis-viton-patten}, our model faithfully reproduces complex and fine-grained clothing patterns. Even for garments with dense textures, irregular motifs, or high-frequency visual elements, the generated results retain clear, sharp, and recognizable patterns with minimal distortion. The strong pattern-preservation ability highlights the effectiveness of our approach in capturing both global appearance and subtle local details.

\paragraph{Complex Structures (VITON-HD).}
Fig.~\ref{fig:supp-vis-viton-structure} further illustrates that our method handles garments with challenging structural designs, such as multi-layered regions, unique silhouettes, or uncommon shapes. The generated try-on results maintain correct garment geometry, coherent contours, and physically plausible spatial arrangements. This demonstrates that our framework models structural priors robustly, enabling accurate synthesis even under significant variations in shape.

\paragraph{DressCode Upper-body, Lower-body, and Dress Sub-sets.}
As shown in Figs.~\ref {fig:supp-vis-upper_body}, \ref {fig:supp-vis-lower_body}, and \ref {fig:supp-vis-dress}, our approach performs consistently well across the three DressCode subsets. In the unpaired and mask-free settings, our model successfully preserves fabric materials, shading, and texture characteristics while achieving realistic alignment between the garment and human body. Across diverse clothing types—including tops, pants, skirts, and full-body dresses—the synthesized results maintain stable structure, smooth boundaries, and visually coherent integration, demonstrating strong generalization and robustness.

\paragraph{In-the-Wild results.}
In addition to controlled benchmark evaluations, RefTON demonstrates strong robustness and generalization in challenging \textit{in-the-wild} scenarios. As shown in Fig.~\ref{fig:in-the-wild}, it produces high-quality try-on results under diverse poses, lighting conditions, and cluttered backgrounds. Our \textbf{mask-free} pipeline directly transfers garments without human parsing masks or pose estimators, while preserving body pose, global structure, and identity cues. Moreover, \textbf{additional-reference try-on} further improves garment geometry, texture details, and overall realism. RefTON also maintains strong background consistency and avoids unnecessary changes to non-garment regions, making it reliable for real-world applications.


%% file: tables/table_dresscode_detail.tex
\begin{table*}[htbp]
  \caption{Quantitative results on three subsets of the DressCode dataset~\cite{morelli2022dress}: upper body, lower body, and dresses. 
  The best and second-best results are highlighted in \textbf{bold} and \underline{underline}, respectively. The symbol ``*'' denotes results reported in prior work, while ``+R'' indicates results with reference image input, and ``MF'' refers to mask-free input images. The subscripts $p$ and $s$ denote specific evaluation metrics for precision and recall, respectively.}
  \label{tab:dresscode-detail}
  \centering
  \resizebox{\linewidth}{!}{
  \begin{tabular}{l*{12}{c}}
    \toprule
    \multirow{2}{*}{\textbf{Method}} 
      & \multicolumn{4}{c}{\textbf{Upper-body}} 
      & \multicolumn{4}{c}{\textbf{Lower-body}} 
      & \multicolumn{4}{c}{\textbf{Dresses}} \\
    \cmidrule(lr){2-5} \cmidrule(lr){6-9} \cmidrule(lr){10-13}
      & \textbf{FID}$_p$$\downarrow$ & \textbf{KID}$_p$$\downarrow$ 
      & \textbf{FID}$_u$$\downarrow$ & \textbf{KID}$_u$$\downarrow$
      & \textbf{FID}$_p$$\downarrow$ & \textbf{KID}$_p$$\downarrow$ 
      & \textbf{FID}$_u$$\downarrow$ & \textbf{KID}$_u$$\downarrow$
      & \textbf{FID}$_p$$\downarrow$ & \textbf{KID}$_p$$\downarrow$ 
      & \textbf{FID}$_u$$\downarrow$ & \textbf{KID}$_u$$\downarrow$ \\
    \midrule
    CAT-DM\(^*\) \cite{zeng2024cat}    
      & 9.85 & 2.38 & 12.62 & 1.89
      & 10.25 & 1.81 & 14.83 & 2.82 
      & 10.71 & 2.02 & 14.30 & 3.36 \\
    OOTDiffusion\(^*\)  ~\cite{xu2025ootdiffusion}
      & 11.03 & 0.29 & -- & --
      & 9.72 & 0.64 & -- & --
      & 10.65 & 0.54 & -- & -- \\
    PromptDresser\(^*\)  \cite{kim2024promptdresser}
      & 11.00 & 0.74 & -- & --
      & 12.55 & 1.46 & -- & --
      & 11.09 & 1.10 & -- & -- \\
    \midrule
    \rowcolor{lightgreen}
    \textbf{RefTON}(Ours)
      & 7.62    & \underline{1.10 }   & \underline{11.13}   & \underline{0.98}
      & \underline{7.60} & 1.38 & 13.07   & 2.11 
      & 7.32  & 1.30 & 11.56    & 1.98    \\
    \rowcolor{lightgreen}
    \textbf{RefTON+R}(Ours)
        & \textbf{6.39}  & \textbf{0.85}  & \textbf{11.08}  & \textbf{0.87} 
        & \textbf{6.61}  & \textbf{1.05}  & \underline{12.56}  & \textbf{1.67} 
        & \textbf{6.09}  & \textbf{1.16}  & 11.16  & 1.72  \\  
    \midrule
    \rowcolor{lightgreen}
    \textbf{RefTON/MF}(Ours)
      & 8.37    & 1.43  & 11.20 & 1.11
      & 8.79   & 1.51   & \textbf{12.50} & \underline{1.83}
      & 7.36  & 1.33   & \underline{10.73} &\underline{1.41} \\
    \rowcolor{lightgreen}
    \textbf{RefTON+R/MF}(Ours)
        & \underline{7.20}   & 1.18  & 11.53 & 1.12
      & 7.85  & \underline{1.21}  & 12.74 & 2.05
      & \underline{6.24 } & \underline{1.20}  & \textbf{10.05} & \textbf{1.30} \\
    \bottomrule
  \end{tabular}}

\end{table*}

%% file: tables/table_merge_appendix.tex
\begin{table*}[htbp]
\caption{
Quantitative comparison across VITON-HD~\cite{choi2021viton} and DressCode~\cite{morelli2022dress} at a resolution of 1024.
The best and second best results are shown in \textbf{bold} and \underline{underline}.
``+R'' denotes the use of reference images, and ``MF'' indicates mask-free inputs. Subscripts $p$ and $u$ represent the \textit{paired} and \textit{unpaired} test settings, respectively. Unless otherwise specified, the same notations carry the same meanings throughout the figures and tables in this paper.
}
\label{tab:combined_reftron_appendix}
\centering
\resizebox{\textwidth}{!}{
\begin{tabular}{l c c c c c c  c c c c c c}
\toprule
\multirow{2}{*}{\textbf{Method}} 
& \multicolumn{6}{c}{\textbf{VITON-HD}} 
& \multicolumn{6}{c}{\textbf{DressCode}} 
\\
\cline{2-13}
& \textbf{LPIPS}$_p$$\downarrow$
& \textbf{SSIM}$_p$$\uparrow$
& \textbf{FID}$_p$$\downarrow$
& \textbf{KID}$_p$$\downarrow$
& \textbf{FID}$_u$$\downarrow$
& \textbf{KID}$_u$$\downarrow$
& \textbf{LPIPS}$_p$$\downarrow$
& \textbf{SSIM}$_p$$\uparrow$
& \textbf{FID}$_p$$\downarrow$
& \textbf{KID}$_p$$\downarrow$
& \textbf{FID}$_u$$\downarrow$
& \textbf{KID}$_u$$\downarrow$
\\
\midrule
\multicolumn{13}{c}{\textit{Mask-based setting}} \\
\midrule
\textbf{RefTON (Ours)}
& \underline{0.079} & \underline{0.870} & \underline{5.96} & \underline{1.05} & \textbf{8.91} & \textbf{1.15}
& \underline{0.056} & \underline{0.899} & \underline{3.28} & \underline{0.76} & \underline{4.84} & \underline{0.83}
\\
\textbf{RefTON+R (Ours)}
& \textbf{0.072} & \textbf{0.873} & \textbf{5.25} & \textbf{0.97} & \underline{9.10} & \underline{1.41}
& \textbf{0.052} & \textbf{0.902} & \textbf{2.84} & \textbf{0.65} & \textbf{4.73} & \textbf{0.76}
\\
\midrule
\multicolumn{13}{c}{\textit{Mask-Free setting}} \\
\midrule
\textbf{RefTON/MF (Ours)}
& \underline{0.068} & \textbf{0.880} & \underline{5.02} & \underline{0.85} & \textbf{8.87} & \textbf{1.05}
& \textbf{0.028} & \textbf{0.956} & \textbf{1.03} & \textbf{0.19} & \textbf{4.24} & \textbf{0.59}
\\
\textbf{RefTON+R/MF (Ours)}
& \textbf{0.067} & \underline{0.875} & \textbf{4.73} & \textbf{0.71} & \underline{8.98} & \underline{1.22}
& \underline{0.030} & \underline{0.953} & \underline{1.15} & \underline{0.25} & \underline{4.41} & \underline{0.69}
\\
\bottomrule
\end{tabular}}
\end{table*}

%% file: main.bib
@String(CVPR= {IEEE Conf. Comput. Vis. Pattern Recog.})

@String(ECCV= {Eur. Conf. Comput. Vis.})

@String(ICLR = {Int. Conf. Learn. Represent.})

@String(AAAI = {AAAI})

@String(CVPR  = {CVPR})

@String(ECCV  = {ECCV})

@String(ICLR  = {ICLR})

@article{binkowski2018demystifying,
  title={Demystifying mmd gans},
  author={Bi{\'n}kowski, Miko{\l}aj and Sutherland, Danica J and Arbel, Michael and Gretton, Arthur},
  journal={arXiv preprint arXiv:1801.01401},
  year={2018}
}

@misc{Seitzer2020FID,
  author={Maximilian Seitzer},
  title={{pytorch-fid: FID Score for PyTorch}},
  month={August},
  year={2020},
  note={Version 0.3.0},
  howpublished={\url{https://github.com/mseitzer/pytorch-fid}},
}

@inproceedings{zhang2018unreasonable,
  title={The unreasonable effectiveness of deep features as a perceptual metric},
  author={Zhang, Richard and Isola, Phillip and Efros, Alexei A and Shechtman, Eli and Wang, Oliver},
  booktitle={Proceedings of the IEEE conference on computer vision and pattern recognition},
  pages={586--595},
  year={2018}
}

@article{nilsson2020understanding,
  title={Understanding ssim},
  author={Nilsson, Jim and Akenine-M{\"o}ller, Tomas},
  journal={arXiv preprint arXiv:2006.13846},
  year={2020}
}

@article{fang2024vivid,
  title={Vivid: Video virtual try-on using diffusion models},
  author={Fang, Zixun and Zhai, Wei and Su, Aimin and Song, Hongliang and Zhu, Kai and Wang, Mao and Chen, Yu and Liu, Zhiheng and Cao, Yang and Zha, Zheng-Jun},
  journal={arXiv preprint arXiv:2405.11794},
  year={2024}
}

@article{feng2025omnitry,
  title={OmniTry: Virtual Try-On Anything without Masks},
  author={Feng, Yutong and Zhang, Linlin and Cao, Hengyuan and Chen, Yiming and Feng, Xiaoduan and Cao, Jian and Wu, Yuxiong and Wang, Bin},
  journal={arXiv preprint arXiv:2508.13632},
  year={2025}
}

@inproceedings{xiao2025omnigen,
  title={Omnigen: Unified image generation},
  author={Xiao, Shitao and Wang, Yueze and Zhou, Junjie and Yuan, Huaying and Xing, Xingrun and Yan, Ruiran and Li, Chaofan and Wang, Shuting and Huang, Tiejun and Liu, Zheng},
  booktitle={Proceedings of the Computer Vision and Pattern Recognition Conference},
  pages={13294--13304},
  year={2025}
}

@inproceedings{guo2025any2anytryon,
  title={Any2anytryon: Leveraging adaptive position embeddings for versatile virtual clothing tasks},
  author={Guo, Hailong and Zeng, Bohan and Song, Yiren and Zhang, Wentao and Liu, Jiaming and Zhang, Chuang},
  booktitle={Proceedings of the IEEE/CVF International Conference on Computer Vision},
  pages={19085--19096},
  year={2025}
}

@inproceedings{choi2024improving,
  title={Improving diffusion models for authentic virtual try-on in the wild},
  author={Choi, Yisol and Kwak, Sangkyung and Lee, Kyungmin and Choi, Hyungwon and Shin, Jinwoo},
  booktitle={European Conference on Computer Vision},
  pages={206--235},
  year={2024},
  organization={Springer}
}

@inproceedings{sohl2015deep,
  title={Deep unsupervised learning using nonequilibrium thermodynamics},
  author={Sohl-Dickstein, Jascha and Weiss, Eric and Maheswaranathan, Niru and Ganguli, Surya},
  booktitle={International conference on machine learning},
  pages={2256--2265},
  year={2015},
  organization={PMLR}
}

@inproceedings{
song2020score,
title={Score-Based Generative Modeling through Stochastic Differential Equations},
author={Yang Song and Jascha Sohl-Dickstein and Diederik P Kingma and Abhishek Kumar and Stefano Ermon and Ben Poole},
booktitle={International Conference on Learning Representations},
year={2021},
url={https://openreview.net/forum?id=PxTIG12RRHS}
}

@inproceedings{kirillov2023segment,
  title={Segment anything},
  author={Kirillov, Alexander and Mintun, Eric and Ravi, Nikhila and Mao, Hanzi and Rolland, Chloe and Gustafson, Laura and Xiao, Tete and Whitehead, Spencer and Berg, Alexander C and Lo, Wan-Yen and others},
  booktitle={Proceedings of the IEEE/CVF international conference on computer vision},
  pages={4015--4026},
  year={2023}
}

@article{Ho2020DenoisingDP,
  title={Denoising Diffusion Probabilistic Models},
  author={Jonathan Ho and Ajay Jain and P. Abbeel},
  journal={ArXiv},
  year={2020},
  volume={abs/2006.11239},
  url={https://api.semanticscholar.org/CorpusID:219955663}
}

@article{Rombach2021HighResolutionIS,
  title={High-Resolution Image Synthesis with Latent Diffusion Models},
  author={Robin Rombach and A. Blattmann and Dominik Lorenz and Patrick Esser and Bj{\"o}rn Ommer},
  journal={2022 IEEE/CVF Conference on Computer Vision and Pattern Recognition (CVPR)},
  year={2021},
  pages={10674-10685},
  url={https://api.semanticscholar.org/CorpusID:245335280}
}

@inproceedings{Chen2018NeuralOD,
  title={Neural Ordinary Differential Equations},
  author={Tian Qi Chen and Yulia Rubanova and Jesse Bettencourt and David Kristjanson Duvenaud},
  booktitle={Neural Information Processing Systems},
  year={2018},
  url={https://api.semanticscholar.org/CorpusID:49310446}
}

@article{Lipman2022FlowMF,
  title={Flow Matching for Generative Modeling},
  author={Yaron Lipman and Ricky T. Q. Chen and Heli Ben-Hamu and Maximilian Nickel and Matt Le},
  journal={ArXiv},
  year={2022},
  volume={abs/2210.02747},
  url={https://api.semanticscholar.org/CorpusID:252734897}
}

@inproceedings{kim2024stableviton,
  title={Stableviton: Learning semantic correspondence with latent diffusion model for virtual try-on},
  author={Kim, Jeongho and Gu, Guojung and Park, Minho and Park, Sunghyun and Choo, Jaegul},
  booktitle={Proceedings of the IEEE/CVF conference on computer vision and pattern recognition},
  pages={8176--8185},
  year={2024}
}

@inproceedings{zhu2023tryondiffusion,
  title={Tryondiffusion: A tale of two unets},
  author={Zhu, Luyang and Yang, Dawei and Zhu, Tyler and Reda, Fitsum and Chan, William and Saharia, Chitwan and Norouzi, Mohammad and Kemelmacher-Shlizerman, Ira},
  booktitle={Proceedings of the IEEE/CVF conference on computer vision and pattern recognition},
  pages={4606--4615},
  year={2023}
}

@inproceedings{choi2021viton,
  title={Viton-hd: High-resolution virtual try-on via misalignment-aware normalization},
  author={Choi, Seunghwan and Park, Sunghyun and Lee, Minsoo and Choo, Jaegul},
  booktitle={Proceedings of the IEEE/CVF conference on computer vision and pattern recognition},
  pages={14131--14140},
  year={2021}
}

@inproceedings{morelli2023ladi,
  title={Ladi-vton: Latent diffusion textual-inversion enhanced virtual try-on},
  author={Morelli, Davide and Baldrati, Alberto and Cartella, Giuseppe and Cornia, Marcella and Bertini, Marco and Cucchiara, Rita},
  booktitle={Proceedings of the 31st ACM international conference on multimedia},
  pages={8580--8589},
  year={2023}
}

@inproceedings{kim2022diffusionclip,
  title={Diffusionclip: Text-guided diffusion models for robust image manipulation},
  author={Kim, Gwanghyun and Kwon, Taesung and Ye, Jong Chul},
  booktitle={Proceedings of the IEEE/CVF conference on computer vision and pattern recognition},
  pages={2426--2435},
  year={2022}
}

@inproceedings{gou2023taming,
  title={Taming the power of diffusion models for high-quality virtual try-on with appearance flow},
  author={Gou, Junhong and Sun, Siyu and Zhang, Jianfu and Si, Jianlou and Qian, Chen and Zhang, Liqing},
  booktitle={Proceedings of the 31st ACM International Conference on Multimedia},
  pages={7599--7607},
  year={2023}
}

@inproceedings{peebles2023scalable,
  title={Scalable diffusion models with transformers},
  author={Peebles, William and Xie, Saining},
  booktitle={Proceedings of the IEEE/CVF international conference on computer vision},
  pages={4195--4205},
  year={2023}
}

@article{goodfellow2020generative,
  title={Generative adversarial networks},
  author={Goodfellow, Ian and Pouget-Abadie, Jean and Mirza, Mehdi and Xu, Bing and Warde-Farley, David and Ozair, Sherjil and Courville, Aaron and Bengio, Yoshua},
  journal={Communications of the ACM},
  volume={63},
  number={11},
  pages={139--144},
  year={2020},
  publisher={ACM New York, NY, USA}
}

@inproceedings{han2018viton,
  title={Viton: An image-based virtual try-on network},
  author={Han, Xintong and Wu, Zuxuan and Wu, Zhe and Yu, Ruichi and Davis, Larry S},
  booktitle={Proceedings of the IEEE conference on computer vision and pattern recognition},
  pages={7543--7552},
  year={2018}
}

@inproceedings{wang2018toward,
  title={Toward characteristic-preserving image-based virtual try-on network},
  author={Wang, Bochao and Zheng, Huabin and Liang, Xiaodan and Chen, Yimin and Lin, Liang and Yang, Meng},
  booktitle={Proceedings of the European conference on computer vision (ECCV)},
  pages={589--604},
  year={2018}
}

@inproceedings{chen2024wear,
  title={Wear-any-way: Manipulable virtual try-on via sparse correspondence alignment},
  author={Chen, Mengting and Chen, Xi and Zhai, Zhonghua and Ju, Chen and Hong, Xuewen and Lan, Jinsong and Xiao, Shuai},
  booktitle={European Conference on Computer Vision},
  pages={124--142},
  year={2024},
  organization={Springer}
}

@inproceedings{xu2025ootdiffusion,
  title={OOTDiffusion: Outfitting Fusion based Latent Diffusion for Controllable Virtual Try-on},
  author={Yuhao Xu and Tao Gu and Weifeng Chen and Chengcai Chen},
  booktitle={AAAI Conference on Artificial Intelligence},
  year={2024},
  url={https://api.semanticscholar.org/CorpusID:268247604}
}

@inproceedings{xu2024magicanimate,
  title={Magicanimate: Temporally consistent human image animation using diffusion model},
  author={Xu, Zhongcong and Zhang, Jianfeng and Liew, Jun Hao and Yan, Hanshu and Liu, Jia-Wei and Zhang, Chenxu and Feng, Jiashi and Shou, Mike Zheng},
  booktitle={Proceedings of the IEEE/CVF Conference on Computer Vision and Pattern Recognition},
  pages={1481--1490},
  year={2024}
}

@article{kingma2013auto,
  title={Auto-encoding variational bayes},
  author={Kingma, Diederik P and Welling, Max},
  journal={arXiv preprint arXiv:1312.6114},
  year={2013}
}

@article{dosovitskiy2020image,
  title={An image is worth 16x16 words: Transformers for image recognition at scale},
  author={Dosovitskiy, Alexey and Beyer, Lucas and Kolesnikov, Alexander and Weissenborn, Dirk and Zhai, Xiaohua and Unterthiner, Thomas and Dehghani, Mostafa and Minderer, Matthias and Heigold, Georg and Gelly, Sylvain and others},
  journal={arXiv preprint arXiv:2010.11929},
  year={2020}
}

@misc{labs2025flux1kontextflowmatching,
      title={FLUX.1 Kontext: Flow Matching for In-Context Image Generation and Editing in Latent Space},
      author={Black Forest Labs and Stephen Batifol and Andreas Blattmann and Frederic Boesel and Saksham Consul and Cyril Diagne and Tim Dockhorn and Jack English and Zion English and Patrick Esser and Sumith Kulal and Kyle Lacey and Yam Levi and Cheng Li and Dominik Lorenz and Jonas Müller and Dustin Podell and Robin Rombach and Harry Saini and Axel Sauer and Luke Smith},
      year={2025},
      eprint={2506.15742},
      archivePrefix={arXiv},
      primaryClass={cs.GR},
      url={https://arxiv.org/abs/2506.15742},
}

@misc{flux2024,
    author={Black Forest Labs},
    title={FLUX},
    year={2024},
    howpublished={\url{https://github.com/black-forest-labs/flux}},
}

@inproceedings{zeng2024cat,
  title={Cat-dm: Controllable accelerated virtual try-on with diffusion model},
  author={Zeng, Jianhao and Song, Dan and Nie, Weizhi and Tian, Hongshuo and Wang, Tongtong and Liu, An-An},
  booktitle={Proceedings of the IEEE/CVF conference on computer vision and pattern recognition},
  pages={8372--8382},
  year={2024}
}

@inproceedings{morelli2022dress,
  title={Dress code: High-resolution multi-category virtual try-on},
  author={Morelli, Davide and Fincato, Matteo and Cornia, Marcella and Landi, Federico and Cesari, Fabio and Cucchiara, Rita},
  booktitle={Proceedings of the IEEE/CVF conference on computer vision and pattern recognition},
  pages={2231--2235},
  year={2022}
}

@article{hu2022lora,
  title={Lora: Low-rank adaptation of large language models.},
  author={Hu, Edward J and Shen, Yelong and Wallis, Phillip and Allen-Zhu, Zeyuan and Li, Yuanzhi and Wang, Shean and Wang, Lu and Chen, Weizhu and others},
  journal={ICLR},
  volume={1},
  number={2},
  pages={3},
  year={2022}
}

@article{su2024roformer,
  title={Roformer: Enhanced transformer with rotary position embedding},
  author={Su, Jianlin and Ahmed, Murtadha and Lu, Yu and Pan, Shengfeng and Bo, Wen and Liu, Yunfeng},
  journal={Neurocomputing},
  volume={568},
  pages={127063},
  year={2024},
  publisher={Elsevier}
}

@inproceedings{shen2025imagdressing,
  title={IMAGDressing-v1: Customizable Virtual Dressing},
  author={Fei Shen and Xin Jiang and Xin He and Hu Ye and Cong Wang and Xiaoyu Du and Zechao Li and Jinghui Tang},
  booktitle={AAAI Conference on Artificial Intelligence},
  year={2024},
  url={https://api.semanticscholar.org/CorpusID:271244829}
}

@article{bai2025qwen2,
  title={Qwen2. 5-vl technical report},
  author={Bai, Shuai and Chen, Keqin and Liu, Xuejing and Wang, Jialin and Ge, Wenbin and Song, Sibo and Dang, Kai and Wang, Peng and Wang, Shijie and Tang, Jun and others},
  journal={arXiv preprint arXiv:2502.13923},
  year={2025}
}

@inproceedings{zheng2019virtually,
  title={Virtually trying on new clothing with arbitrary poses},
  author={Zheng, Na and Song, Xuemeng and Chen, Zhaozheng and Hu, Linmei and Cao, Da and Nie, Liqiang},
  booktitle={Proceedings of the 27th ACM international conference on multimedia},
  pages={266--274},
  year={2019}
}

@article{chong2024catvton,
  title={Catvton: Concatenation is all you need for virtual try-on with diffusion models},
  author={Chong, Zheng and Dong, Xiao and Li, Haoxiang and Zhang, Shiyue and Zhang, Wenqing and Zhang, Xujie and Zhao, Hanqing and Jiang, Dongmei and Liang, Xiaodan},
  journal={arXiv preprint arXiv:2407.15886},
  year={2024}
}

@article{chong2025catv2ton,
  title={Catv2ton: Taming diffusion transformers for vision-based virtual try-on with temporal concatenation},
  author={Chong, Zheng and Zhang, Wenqing and Zhang, Shiyue and Zheng, Jun and Dong, Xiao and Li, Haoxiang and Wu, Yiling and Jiang, Dongmei and Liang, Xiaodan},
  journal={arXiv preprint arXiv:2501.11325},
  year={2025}
}

@article{yang2025omnivton,
  title={OmniVTON: Training-Free Universal Virtual Try-On},
  author={Yang, Zhaotong and Li, Yuhui and He, Shengfeng and Li, Xinzhe and Xu, Yangyang and Dong, Junyu and Du, Yong},
  journal={arXiv preprint arXiv:2507.15037},
  year={2025}
}

@article{kim2024promptdresser,
  title={PromptDresser: Improving the Quality and Controllability of Virtual Try-On via Generative Textual Prompt and Prompt-aware Mask},
  author={Kim, Jeongho and Jin, Hoiyeong and Park, Sunghyun and Choo, Jaegul},
  journal={arXiv preprint arXiv:2412.16978},
  year={2024}
}

@article{liu2022flow,
  title={Flow straight and fast: Learning to generate and transfer data with rectified flow},
  author={Liu, Xingchao and Gong, Chengyue and Liu, Qiang},
  journal={arXiv preprint arXiv:2209.03003},
  year={2022}
}

@inproceedings{guler2018densepose,
  title={Densepose: Dense human pose estimation in the wild},
  author={G{\"u}ler, R{\i}za Alp and Neverova, Natalia and Kokkinos, Iasonas},
  booktitle={Proceedings of the IEEE conference on computer vision and pattern recognition},
  pages={7297--7306},
  year={2018}
}

@inproceedings{toshev2014deeppose,
  title={Deeppose: Human pose estimation via deep neural networks},
  author={Toshev, Alexander and Szegedy, Christian},
  booktitle={Proceedings of the IEEE conference on computer vision and pattern recognition},
  pages={1653--1660},
  year={2014}
}

@article{openpose,
  author = {Z. {Cao} and G. {Hidalgo Martinez} and T. {Simon} and S. {Wei} and Y. A. {Sheikh}},
  journal = {IEEE Transactions on Pattern Analysis and Machine Intelligence},
  title = {OpenPose: Realtime Multi-Person 2D Pose Estimation using Part Affinity Fields},
  year = {2019}
}

@inproceedings{cao2017realtime,
  author = {Zhe Cao and Tomas Simon and Shih-En Wei and Yaser Sheikh},
  booktitle = {CVPR},
  title = {Realtime Multi-Person 2D Pose Estimation using Part Affinity Fields},
  year = {2017}
}

@inproceedings{wei2016cpm,
  author = {Shih-En Wei and Varun Ramakrishna and Takeo Kanade and Yaser Sheikh},
  booktitle = {CVPR},
  title = {Convolutional pose machines},
  year = {2016}
}

@article{li2020self,
  title={Self-Correction for Human Parsing}, 
  author={Li, Peike and Xu, Yunqiu and Wei, Yunchao and Yang, Yi},
  journal={IEEE Transactions on Pattern Analysis and Machine Intelligence}, 
  year={2020},
  doi={10.1109/TPAMI.2020.3048039}}

@inproceedings{dong2014towards,
  title={Towards unified human parsing and pose estimation},
  author={Dong, Jian and Chen, Qiang and Shen, Xiaohui and Yang, Jianchao and Yan, Shuicheng},
  booktitle={Proceedings of the IEEE Conference on Computer Vision and Pattern Recognition},
  pages={843--850},
  year={2014}
}

@article{ravi2024sam2,
  title={SAM 2: Segment Anything in Images and Videos},
  author={Ravi, Nikhila and Gabeur, Valentin and Hu, Yuan-Ting and Hu, Ronghang and Ryali, Chaitanya and Ma, Tengyu and Khedr, Haitham and R{\"a}dle, Roman and Rolland, Chloe and Gustafson, Laura and Mintun, Eric and Pan, Junting and Alwala, Kalyan Vasudev and Carion, Nicolas and Wu, Chao-Yuan and Girshick, Ross and Doll{\'a}r, Piotr and Feichtenhofer, Christoph},
  journal={arXiv preprint arXiv:2408.00714},
  url={https://arxiv.org/abs/2408.00714},
  year={2024}
}

@article{he2024wildvidfit,
  title={WildVidFit: Video Virtual Try-On in the Wild via Image-Based Controlled Diffusion Models},
  author={He, Zijian and Chen, Peixin and Wang, Guangrun and Li, Guanbin and Torr, Philip HS and Lin, Liang},
  journal={arXiv preprint arXiv:2407.10625},
  year={2024}
}

@article{meng2025hf,
  title={HF-VTON: High-Fidelity Virtual Try-On via Consistent Geometric and Semantic Alignment},
  author={Meng, Ming and Dong, Qi and Li, Jiajie and Zhu, Zhe and Wang, Xingyu and Fan, Zhaoxin and Zhao, Wei and Wu, Wenjun},
  journal={arXiv preprint arXiv:2505.19638},
  year={2025}
}

@article{dinh2016density,
  title={Density estimation using real nvp},
  author={Dinh, Laurent and Sohl-Dickstein, Jascha and Bengio, Samy},
  journal={arXiv preprint arXiv:1605.08803},
  year={2016}
}

@article{dinh2014nice,
  title={Nice: Non-linear independent components estimation},
  author={Dinh, Laurent and Krueger, David and Bengio, Yoshua},
  journal={arXiv preprint arXiv:1410.8516},
  year={2014}
}

@article{kingma2018glow,
  title={Glow: Generative flow with invertible 1x1 convolutions},
  author={Kingma, Durk P and Dhariwal, Prafulla},
  journal={Advances in neural information processing systems},
  volume={31},
  year={2018}
}

@article{velioglu2025enhancing,
  title={Enhancing Person-to-Person Virtual Try-On with Multi-Garment Virtual Try-Off},
  author={Velioglu, Riza and Bevandic, Petra and Chan, Robin and Hammer, Barbara},
  journal={arXiv preprint arXiv:2504.13078},
  year={2025}
}

@ARTICLE{ViTON-GUN,
  author={Zhang, Nannan and Xie, Zhenyu and Sun, Zhengwentai and Zhu, Hairui and Jin, Zirong and Xiang, Nan and Han, Xiaoguang and Wu, Song},
  journal={IEEE Transactions on Visualization and Computer Graphics}, 
  title={ViTon-GUN: Person-to-Person Virtual Try-on via Garment Unwrapping}, 
  year={2025},
  volume={31},
  number={10},
  pages={7740-7751},
  doi={10.1109/TVCG.2025.3550776}}

@article{cui2023street-tryon,
  title={Street TryOn: Learning In-the-Wild Virtual Try-On from Unpaired Person Images},
  author={Cui, Aiyu and Mahajan, Jay and Shah, Viraj and Gomathinayagam, Preeti and Lazebnik, Svetlana},
  journal={arXiv preprint arXiv:2311.16094},
  year={2023}
}

@article{cheng2024hico,
  title={Hico: Hierarchical controllable diffusion model for layout-to-image generation},
  author={Cheng, Bo and Ma, Yuhang and Wu, Liebucha and Liu, Shanyuan and Ma, Ao and Wu, Xiaoyu and Leng, Dawei and Yin, Yuhui},
  journal={Advances in neural information processing systems},
  volume={37},
  pages={128886--128910},
  year={2024}
}

@inproceedings{he2025plangen,
  title={Plangen: Towards unified layout planning and image generation in auto-regressive vision language models},
  author={He, Runze and Cheng, Bo and Ma, Yuhang and Jia, Qingxiang and Liu, Shanyuan and Ma, Ao and Wu, Xiaoyu and Wu, Liebucha and Leng, Dawei and Yin, Yuhui},
  booktitle={Proceedings of the IEEE/CVF International Conference on Computer Vision},
  pages={18143--18154},
  year={2025}
}

@article{Liu_2025,
   title={Bridge Diffusion Model: Bridge Chinese Text-to-Image Diffusion Model with English Communities},
   volume={39},
   ISSN={2159-5399},
   url={http://dx.doi.org/10.1609/aaai.v39i5.32590},
   DOI={10.1609/aaai.v39i5.32590},
   number={5},
   journal={Proceedings of the AAAI Conference on Artificial Intelligence},
   publisher={Association for the Advancement of Artificial Intelligence (AAAI)},
   author={Liu, Shanyuan and Cheng, Bo and Ma, Yuhang and Wu, Liebucha and Ma, Ao and Wu, Xiaoyu and Leng, Dawei and Yin, Yuhui},
   year={2025},
   month=apr, pages={5541–5549} }

@article{ma2025nami,
  title={NAMI: Efficient Image Generation via Bridged Progressive Rectified Flow Transformers},
  author={Ma, Yuhang and Cheng, Bo and Liu, Shanyuan and Zhou, Hongyi and Wu, Liebucha and Wu, Xiaoyu and Leng, Dawei and Yin, Yuhui},
  journal={arXiv preprint arXiv:2503.09242},
  year={2025}
}

@misc{liu2025nanocontrollightweightframeworkprecise,
      title={NanoControl: A Lightweight Framework for Precise and Efficient Control in Diffusion Transformer}, 
      author={Shanyuan Liu and Jian Zhu and Junda Lu and Yue Gong and Liuzhuozheng Li and Bo Cheng and Yuhang Ma and Liebucha Wu and Xiaoyu Wu and Dawei Leng and Yuhui Yin},
      year={2025},
      eprint={2508.10424},
      archivePrefix={arXiv},
      primaryClass={cs.CV},
      url={https://arxiv.org/abs/2508.10424}, 
}

@article{zhu2025flux,
  title={FLUX-Makeup: High-Fidelity, Identity-Consistent, and Robust Makeup Transfer via Diffusion Transformer},
  author={Zhu, Jian and Liu, Shanyuan and Li, Liuzhuozheng and Gong, Yue and Wang, He and Cheng, Bo and Ma, Yuhang and Wu, Liebucha and Wu, Xiaoyu and Leng, Dawei and others},
  journal={arXiv preprint arXiv:2508.05069},
  year={2025}
}

@article{gong2025cta,
  title={CTA-Flux: Integrating Chinese Cultural Semantics into High-Quality English Text-to-Image Communities},
  author={Gong, Yue and Liu, Shanyuan and Li, Liuzhuozheng and Zhu, Jian and Cheng, Bo and Wu, Liebucha and Wu, Xiaoyu and Ma, Yuhang and Leng, Dawei and Yin, Yuhui},
  journal={arXiv preprint arXiv:2508.14405},
  year={2025}
}
